%% file: templatePRIME.tex
\documentclass{article}

\usepackage{PRIMEarxiv}

\usepackage{tikz}
\usetikzlibrary{calc,decorations.pathreplacing,calligraphy,patterns}
\usepackage{tikzlings}

\usepackage[utf8]{inputenc} 
\usepackage[T1]{fontenc}    
\usepackage{url}            
\usepackage{booktabs}       
\usepackage{amsfonts}       
\usepackage{nicefrac}       
\usepackage{microtype}      
\usepackage{lipsum}
\usepackage{graphicx}
\graphicspath{{media/}}     

\usepackage{graphicx}
\usepackage{amsmath}
\usepackage{amssymb}
\usepackage{booktabs}
\usepackage{multirow}
\usepackage[prefix=]{xcolor-material}
\usepackage{colortbl}
\usepackage{adjustbox}
\usepackage{enumitem}

\newcommand{\code}{\texttt}

\pgfmathdeclarerandomlist{material}{%
{Red}{Blue}{Green}}
\tikzset{%
  half clip/.code={
    \clip (0, -256) rectangle (256, 256);
  },
  color/.code=\colorlet{fill color}{#1},
  color alias/.code args={#1 as #2}{\colorlet{#1}{#2}},
  colors alias/.style={color alias/.list/.expanded={#1}},
  execute/.code={#1},
  on left/.style={.. on left/.style={#1}},
  on right/.style={.. on right/.style={#1}},
  split/.style args={#1 and #2}{
    on left ={color alias=fill color as #1},
    on right={color alias=fill color as #2, half clip}
  }
}
\newcommand\reflect[2][]{%
\begin{scope}[#1]\foreach \side in {-1, 1}{\begin{scope}
\ifnum\side=-1 \tikzset{.. on left/.try}\else\tikzset{.. on right/.try}\fi
\begin{scope}[xscale=\side]#2\end{scope}
\end{scope}}\end{scope}}

\definecolor{Grey100}{HTML}{F5F5F5}
\definecolor{Purple100}{HTML}{E1BEE7}
\definecolor{BlueGrey900}{HTML}{263238}
\definecolor{BlueGrey500}{HTML}{607D8B}
\definecolor{BlueGrey100}{HTML}{CFD8DC}
\definecolor{Yellow50}{HTML}{FFFDE7}
\tikzset{cat/.pic={
      \tikzset{x=3cm/5,y=3cm/5,shift={(0,-1/3)}}
      \useasboundingbox (-1,-1) (1,2);
      \fill [BlueGrey900] (0,-2)
        .. controls ++(180:3) and ++(0:5/4) .. (-2,0)
        arc (270:90:1/5)
        .. controls ++(0:2) and ++(180:11/4) .. (0,-2+2/5);
      \foreach \i in {-1,1}
        \scoped[shift={(1/2*\i,9/4)}, rotate=45*\i]{
          \clip [overlay] (0, 5/9) ellipse [radius=8/9];
          \clip [overlay] (0,-5/9) ellipse [radius=8/9];
          \fill [BlueGrey900] ellipse [radius=1];
          \clip [overlay] (0, 7/9) ellipse [radius=10/11];
          \clip [overlay] (0,-7/9) ellipse [radius=10/11];
          \fill [Purple100] ellipse [radius=1];
        };
      \fill [BlueGrey900] ellipse [x radius=3/4, y radius=2];
      \fill [BlueGrey100] ellipse [x radius=1/3, y radius=1];
      \fill [BlueGrey900]
        (0,15/8) ellipse [x radius=1, y radius=5/6]
        (0, 8/6) ellipse [x radius=1/2, y radius=1/2]
        {[shift={(-1/2,-2)}, rotate= 10]  ellipse [x radius=1/3, y radius=5/4]}
        {[shift={( 1/2,-2)}, rotate=-10] ellipse [x radius=1/3, y radius=5/4]};
      \fill [BlueGrey500]
        (-1/9,11/8) ellipse [x radius=1/5, y radius=1/5]
        ( 1/9,11/8) ellipse [x radius=1/5, y radius=1/5];
      \fill [Purple100]
        (0,12/8)     ellipse [x radius=1/10, y radius=1/5]
        (0,12/8+1/9) ellipse [x radius=1/5 , y radius=1/10];
      \foreach \i in {-1,1}
        \scoped[shift={(1/2*\i,2)}, rotate=35*\i]{
          \clip [overlay] (0, 1/7) ellipse [radius=2/7];
          \clip [overlay] (0,-1/7) ellipse [radius=2/7];
          \fill [Yellow50] ellipse [radius=1];
        };
      \scoped{
        \clip (-1,-2) rectangle ++(2,1);
        \fill [BlueGrey900] (0,-2) ellipse [radius=1/2];
        \fill [Grey100]
          (-1/2,-2) ellipse [x radius=1/3, y radius=1/4]
          ( 1/2,-2) ellipse [x radius=1/3, y radius=1/4];
      };
      \foreach \i in {-1,1}
        \foreach \j in {-1,0,1}
          \fill [Grey100, shift={(0,11/8)}, xscale=\i, rotate=\j*15,
            shift=(0:1/2)]
            ellipse [x radius=1/3, y radius=1/64];
    }}

\tikzset{owl/.pic={
\begin{scope}[x=3cm/512,y=3cm/512]
\foreach \i in {-3,...,3}
  \fill [BlueGrey900, shift={(0,128)}, rotate=-\i*13]
    (0,0) arc (90:450:48 and 144);
\foreach \i in {-1,...,1}
  \fill [BlueGrey900, shift={(0,32)}, rotate=-\i*10]
    (0,0) arc (90:450:32 and 128);
\reflect[split={BlueGrey700 and BlueGrey800}]{
  \fill [fill color] (0,224)
  .. controls ++(180:96) and ++( 90:96) .. (-112,32)
  .. controls ++(270:96) and ++(180: 0) .. (0,-148)
  .. controls ++(  0: 0) and ++(270:96) .. ( 112,32)
  .. controls ++( 90:96) and ++(  0:96) .. cycle;
}
\reflect[
  on left ={colors alias={
    outer eye as BlueGrey600, eyebrow as BlueGrey800, chest as BlueGrey100}},
  on right={colors alias={
    outer eye as BlueGrey700, eyebrow as BlueGrey900, chest as BlueGrey200},
    half clip}]{
  \fill [outer eye]   (64, 128) circle [radius=80];
  \fill [Grey100]     (64, 128) circle [radius=40];
  \fill [BlueGrey900] (64, 128) circle [radius=20];
  \fill [eyebrow]  (0,112)
    .. controls ++( 90:128) and ++(270:64) .. (160,240)
    .. controls ++(270:112) and ++( 90:64) .. cycle;
  \fill [chest] 
    (0,-140)
    .. controls ++(135:32) and ++(270:32) .. (-64,-40)
    arc (180:0:64)
    .. controls ++(270:32) and ++( 45:32) .. cycle;
}
\reflect[
  on left ={colors alias={
    beak as Amber500, perch as BlueGrey800, foot as BlueGrey600, 
    talon as Amber500}},
  on right={colors alias={
    beak as Amber700, perch as BlueGrey900, foot as BlueGrey700, 
      talon as Amber700}, half clip}]{
  \fill [beak] (0,28)
    .. controls ++(135:32) and ++(270:16) .. (-32,64)
    arc (180:0:32)
    .. controls ++(270:16) and ++(45:32) .. cycle;
  \fill [perch]
    (-192,-128) arc (90:270:12) -- ++(384,0) arc (270:450:12) -- cycle;
  \tikzset{shift={(0, -128)}}
  \fill [foot]
    (3, 0) arc (180:0:36 and 24) -- cycle;
  \foreach \j in {-1,0,1}
    \scoped[shift={(40+20*\j,0)}]\fill [talon]
       (10,0) arc (0:180:10) -- (0,-16) -- cycle;
}
\end{scope}},}

\newcommand*{\inputsize}{\ensuremath{I_{\text{res}}}}
\newcommand*{\layer}{\ensuremath{\ell}}

\newcommand*{\rfsize}[1][\layer]{\ensuremath{r_{#1}}}
\newcommand*{\rfsizemin}[1][\layer]{\rfsize[#1,\text{min}]}
\newcommand*{\rfsizemax}[1][\layer]{\rfsize[#1,\text{max}]}

\newcommand*{\kernelsize}[1][\layer]{\ensuremath{k_{#1}}}
\newcommand*{\stridesize}[1][\layer]{\ensuremath{g_{#1}}}

\usepackage[pagebackref,breaklinks,colorlinks]{hyperref}
\hypersetup{
    citecolor={purple},
    linkcolor={blue}
}

\usepackage[capitalize]{cleveref}
\crefname{section}{Sec.}{Secs.}
\Crefname{section}{Section}{Sections}
\Crefname{table}{Table}{Tables}
\crefname{table}{Tab.}{Tabs.}

\def\cvprPaperID{11925} 
\def\confName{CVPR}
\def\confYear{2023}

\title{Receptive Field  Refinement for Convolutional Neural Networks Reliably Improves Predictive Performance
}

\author{
  Mats L. Richter  \\
  Mila - Quebec AI Institute \\
  Université de Montréal \\
  Montreal\\
  \texttt{mats-leon.richter@mila.quebec} \\
   \And
  Christopher Pal \\
  Mila - Quebec AI Institute \\
  Polytechnique Montreal \\
  Canada CIFAR AI Chair \\
  \texttt{christopher.pal@polymtl.ca} \\
}

\begin{document}
\maketitle

\begin{abstract}
    Minimal changes to neural architectures (e.g. changing a single hyperparameter in a key layer), can lead to significant gains in predictive performance in Convolutional Neural Networks (CNNs).
    In this work, we present a new approach to receptive field analysis that can yield these types of theoretical and empirical performance gains across twenty well-known CNN architectures examined in our experiments.
    By further developing and formalizing the analysis of receptive field expansion in convolutional neural networks, we can predict unproductive layers in an automated manner before ever training a model. 
    This allows us to optimize the parameter-efficiency of a given architecture at low cost.
    Our method is computationally simple and can be done in an automated manner or even manually with minimal effort for most common architectures.
    We demonstrate the effectiveness of this approach by increasing parameter efficiency across past and current top-performing CNN-architectures.
    Specifically, our approach is able to improve ImageNet1K performance across a wide range of well-known, state-of-the-art (SOTA) model classes, including: 
    VGG Nets, MobileNetV1, MobileNetV3, NASNet A (mobile), MnasNet, EfficientNet, and ConvNeXt -- \textbf{leading to a new SOTA result for each model class}.  
    
\end{abstract}


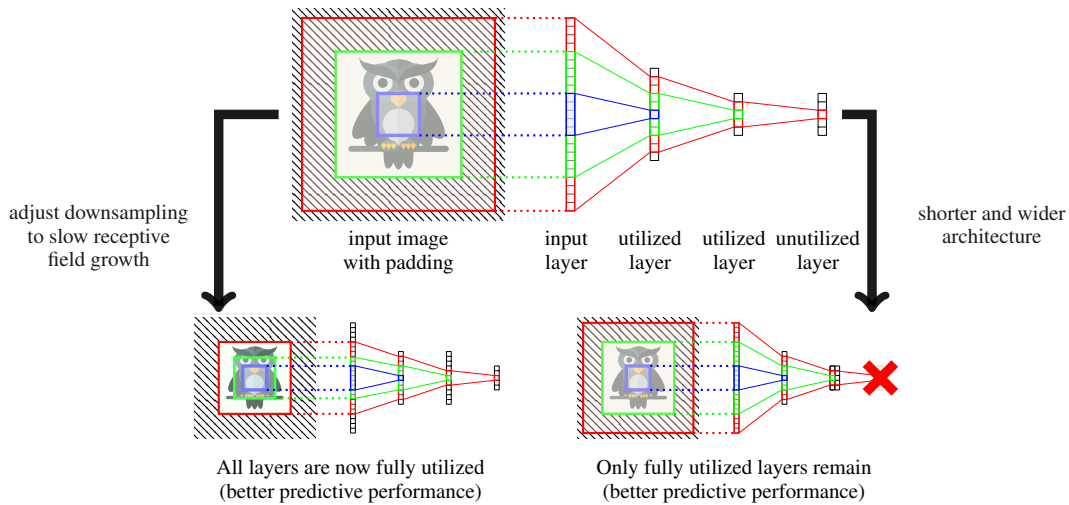
\begin{figure*}[htb!]
  \centering
  \begin{adjustbox}{max size={.9\textwidth}{.9\textheight}}
  \input{latex/fi-opt}
  \end{adjustbox}
  \caption{\textbf{Receptive-field-based refining of a CNN:} Layers with oversized receptive fields are underutilized and therefore inefficient. Slowing the receptive field (RF) growth reactivates these layers, increasing predictive performance. Alternatively, such layers can be removed, and the network width can be adjusted to account for the cut parameters.}
    \label{fig:overview}
\end{figure*}

\section{Introduction}
\label{sec:intro}


Deep learning in computer vision has shown a long-term trend towards bigger and computationally more expensive architectures to achieve higher predictive performance \cite{amoebanet, nfnet}.
Considering that not every developer has access to the compute resources necessary for training SOTA-models, the current state of trial-and-error-heavy development becomes increasingly unsustainable.
Unfortunately, modern Neural Architecture Search (NAS) algorithms suffer from similar problems, since they require an extensive number of costly model-training cycles to yield competitive results (e.g. on standard ImageNet1k evaluations) \cite{mnasnet, nasnet, amoebanet}. 

In contrast to this, we develop a methodology for improving the predictive performance of convolutional neural network (CNN) architectures \textbf{reliably and cheaply}.
This is achieved by analyzing the receptive field of the individual layers and using this information to adapt the model to the input resolution, removing or reactivating underutilized layers in the process.

To demonstrate this, we apply minor changes derived from our analysis to a variety of popular architectures (like MobileNetV3, EfficientNet, and ConvNeXt \cite{mobilenetv3, efficientnet, convnext}). Our experiments show that model changes derived through our approach consistently yield significant performance improvements.
Importantly, our approach is based only on neural architecture properties and the resolution of input imagery. 
Therefore, a model does not need to be trained or even initialized to diagnose network inefficiencies, reducing trial-and-error greatly.
Our proposed procedure is also different from zero-cost NAS, \cite{zerocostnas}, which can be summarized as forms of low-cost proxies for the performance of models relative to each other.
Instead, we identify and localize inefficiencies inside a specific architecture and demonstrate how to resolve these, gaining reliable increases in predictive performance.
In short, our core contributions are as follows:

\setlist{nolistsep}
\begin{itemize}[noitemsep]
    \item We propose a simple analysis method, the ``minimal feasible input resolution ($I_{min}$)'', which can be used to quickly and simply uncover design inefficiencies in CNN architectures (section \ref{sec:imin}).
    \item Using $I_{min}$, we uncover design inefficiencies across twenty popular CNN models, even in architectures specifically designed to be efficient such as MobileNetV3 \cite{mobilenetv3} and EfficientNet \cite{efficientnet} (section \ref{sec:flaws}).
    \item Based on our method, we propose general strategies for refining these inefficient architectures to increase predictive performance (section \ref{sec:flaws}). We also provide an algorithm that automates the proposed analysis in our supplementary material.
    \item We resolve the discovered inefficiencies with minimal changes to the CNNs, reliably improving predictive performance in all cases (section 
    \ref{sec:effresolving}, section \ref{sec:restresolving}).
\end{itemize}

\section{Related Work}

\begin{figure}[htb]

\label{sec:superstructure}
  \centering
  
  \includegraphics[scale=0.3]{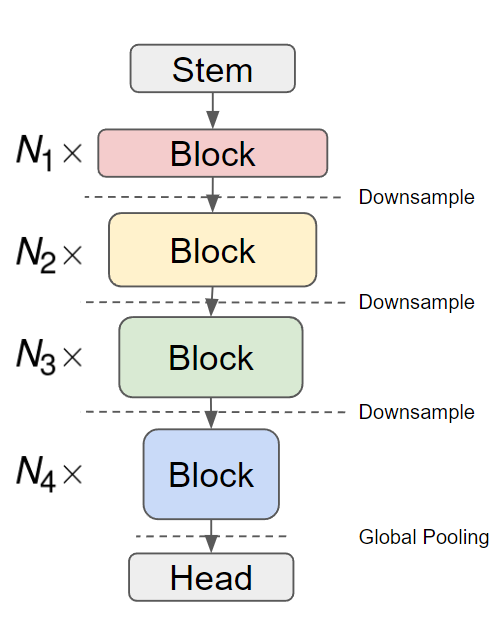}
  \caption{An example for a \textbf{the superstructure of a CNN}, in which the building blocks are embedded.}
    \label{fig:superstructure}
\end{figure}

The overall topologies of modern CNN architectures mainly orient themselves on building-block-style models \cite{efficientnet, efficientnetv2, nfnet} highly driven by convention.
This is, to some degree, necessary, since the search space of neural architectures would otherwise be excessively large \cite{nasnet}. However, the current convention also introduces biases into the chosen design directions, favoring for example the optimization of building blocks over the layout of the overall superstructure of the model \cite{efficientnet, efficientnetv2, mobilenetv2, mobilenetv3, nasnet, mnasnet, amoebanet, resnet, resnet2, wide_resnet}.
The superstructure is the overall layout of the neural architecture, in which the building blocks are embedded (see example in figure \ref{fig:superstructure}).
The main feature extractor is subdivided into multiple stages of building blocks with increasing filter sizes and separated by downsampling layers, followed by the output head of the model.
Our proposed analysis method for refining neural architectures is primarily operating on the level of the superstructure by suggesting changes in the number of layers or the number and positioning of downsampling layers.
Therefore, our approach should be seen not as contrary but as complementary to the current design processes in CNN-architectures. 
This allows users to reliably tap into currently underexplored design-pathways for optimizing the predictive performance of a CNN architecture.

The receptive field is a widely known property of convolutional neural networks and has been studied from various angles.
The works of \cite{araujo2019computing} and \cite{floridaRF} have explored receptive fields from an analytical perspective. The work of \cite{araujo2019computing} in particular provides the basic formal definitions and computational methods that are used throughout this work.
An empirical study on the effective receptive field was conducted by \cite{effRF}, demonstrating that the actual receptive field is substantially smaller than the formal definition suggests.
All the prior works above on receptive fields have suggested that the receptive field could be used to optimize neural architectures, but they did not provide concrete approaches to reliably adjust architectures to improve performance.
%
At the same time, since the receptive field is such a central property of CNNs, it could be argued that it has been leveraged implicitly to improve model performance in many independent cases. 
For example, invariance to object size and problem-specific improvements in tasks like cancer mammography classification \cite{cancer1, cancer2, cancer3} and object detection \cite{fpn, yolov3, yolov4} are commonly achieved by recombining features from layers with heterogeneous receptive field sizes.\\
In contrast to more informal, observational, and implicit procedures,  we analyze receptive fields explicitly to improve the predictive quality of CNN models.
Other concrete attempts to explicitly utilize receptive field analysis to improve existing architectures include \cite{goingdeeper}, who proposed a pruning strategy to improve  predictive performance on low-resolution datasets.
Other work such as DynOPool \cite{rfpooling}, Shape Adaptor \cite{shapeadaptor} and N-Jet \cite{scaletheory} adapt the downsampling rate (and therefore the receptive field size) that is trained alongside the network.
Our approach deviates from all of these in multiple significant ways.
First, DynOPool, Shape Adaptor, and N-Jet learn the downsampling procedures during the training process, whereas we use an offline analysis to guide design decisions made before training.
The aforementioned techniques also alter the model footprint during training, resulting in potential instabilities, making resource consumption harder to control.
In contrast, our approach is a substantially less complex procedure.
It does not require training a model, since it is purely based on an analysis of the architecture, and it can be utilized to deliver consistent improvements on a wide range of models on visually complex datasets like ImageNet1k.
Finally, despite the higher complexity, prior work has not succeeded in demonstrating competitive increases in CNN predictive performance with contemporary and current SOTA-models on standard datasets like ImageNet1K, which we do achieve with our work.



\section{Our Method}
\label{sec:imin}
In this section, we elaborate on the formal and theoretical foundation of our approach. We first introduce the general concept of the receptive field and the minimum receptive field size.
We then briefly discuss unproductive layers in CNN-architectures, a predictable and localizable inefficiency in a CNN caused by a mismatch between input resolutions and CNN-architecture.
Finally, we characterize underutilized CNN-architectures based on these predictable inefficiencies and propose two basic approaches on how to modify the superstructure of an arbitrary feed-forward CNN architecture to resolve them.

\begin{figure}
  \centering
  \begin{adjustbox}{max size={.45\textwidth}{.4\textheight}}
  \begin{tikzpicture}[
    A/.style={red,fill=white!90!red,fill opacity=.3},
    Bmax/.style={blue!70!black,fill=white!90!green,fill opacity=.3},
    Bmin/.style={green!80!white,fill=white!90!green,fill opacity=.3},
    C/.style={blue,fill=white!90!blue,fill opacity=.3},
    ]
    \fill[white!90!black]
      (-2.52,-2.52) coordinate (out1) rectangle (2.52,2.52) coordinate (out2);
    \draw[fill=white]
      (-2.3,-2.3) coordinate (A1) rectangle (2.3,2.3) coordinate (A2) ;
      \pic{owl};
    \draw[Bmax,line width=1.6pt]
      (-1.7,-1.7) coordinate (Bmax1) rectangle (1.7,1.7) coordinate (Bmax2);
    \draw[Bmin,line width=1.6pt]
      (-.5,-.5) coordinate (Bmin1) rectangle (.5,.5) coordinate (Bmin2);
    \node[align=center] at (0,-3.1) {input image};

    \begin{scope}[shift={(4,0)}]
      \coordinate (left0) at (0,0);
      \coordinate (right0) at (.2,0);
      \draw[help lines,step=.2,shift={(0,.1)}] (A1-|left0) grid (A2-|right0);
      \draw[Bmax,line width=1pt]
        (Bmax1-|left0) rectangle (Bmax2-|right0);
      \draw[Bmin,line width=.6pt]
        (Bmin1-|left0) rectangle (Bmin2-|right0);
      \node[align=center] at (0,-3) {input\\layer};
    \end{scope}

    \begin{scope}[shift={(6,0)}]
      \coordinate (left1) at (0,0);
      \coordinate (right1) at (.2,0);
      \draw[help lines,step=.2,shift={(0,.1)}] (0,-.2) grid (0.2,2.0);
      \draw[Bmax,line width=1pt]
        (0,.7) coordinate (Bmax1L1) rectangle (.2,1.3) coordinate (Bmax2L1);

      \draw[help lines,step=.2,shift={(0,.1)}] (0,-2.2) grid (0.2,-.6);
      \draw[Bmin,line width=.6pt]
        (0,-1.5) coordinate (Bmin1L1) rectangle (.2,-1.3) coordinate (Bmin2L1);
      \node[align=center] at (0,-3) {parallel\\layers};
    \end{scope}

    \begin{scope}[shift={(8,0)}]
      \coordinate (left2) at (0,0);
      \coordinate (right2) at (.2,0);
      \draw[help lines,step=.2,shift={(0,.1)}] (0,-1.2) grid (0.2,1.0);
      \draw[Bmax,line width=1pt]
        (0,-.1) coordinate (B1L2) rectangle (.2,.1) coordinate (B2L2);
      \node[align=center] at (0,-3) {combined\\layer};
    \end{scope}

    \draw[Bmax,dotted,line width=1.pt] (Bmax2) -- (Bmax2-|left0)
      (Bmax1-|Bmax2) -- (Bmax1-|left0);
    \draw[Bmin,dotted,line width=1.pt] (Bmin2) -- (Bmin2-|left0)
      (Bmin1-|Bmin2) -- (Bmin1-|left0);

    \draw[Bmax] (Bmax1-|right0) -- (Bmax1L1-|left1)
      (Bmax2-|right0) -- (Bmax2L1-|left1);
    \draw[Bmin] (Bmin1-|right0) -- (Bmin1L1-|left1)
      (Bmin2-|right0) -- (Bmin2L1-|left1);

    \draw[Bmax] (Bmax1L1-|right1) -- (B1L2-|left2)
      (Bmax2L1-|right1) -- (B2L2-|left2);
    \draw[Bmin] (Bmin1L1-|right1) -- (B1L2-|left2)
      (Bmin2L1-|right1) -- (B2L2-|left2);
    
  \end{tikzpicture}
  \end{adjustbox}
  
  \caption{The value for $\rfsizemax[\layer]$ is computed from the path with the \textbf{largest} receptive field (blue). $\rfsizemin[\layer]$ is computed from the path of \textbf{smallest} receptive field (green).}
  \label{fig:catmp}
\end{figure}
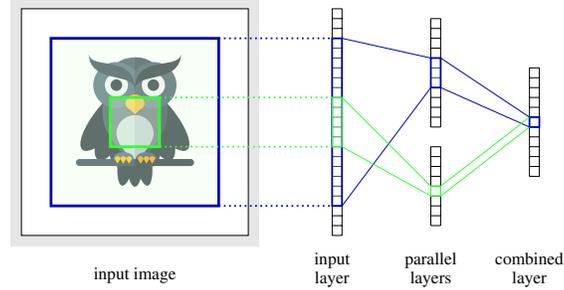

\subsection{The Receptive Field}
For a sequential, fully convolutional network $L$ with layers $\layer{} = 1, 2, 3, ..., L$, the receptive field of a layer can be described as an area of shape $(\rfsize[\layer{}] \times \rfsize[\layer{}])$ on the input image. Only information present in this area can influence the output of a specific feature, locally restricting the feature extraction.
For the first layer, the receptive field is equivalent to the shape of the kernel $\rfsize[0] = \kernelsize[0]$. We are assuming square kernels in all layers, to simplify the computation.
Consecutive layers with a kernel size $(\kernelsize[\layer{}] \times \kernelsize[\layer{}]) > (1 \times 1)$ expand the receptive field by covering multiple input-feature-map positions. Essentially, $\rfsize[\layer{}]$ grows by $\kernelsize[\layer{}]-1$ from the previous layer.
Downsampling accelerates this process further, by reducing the resolution of the input feature map of the following layers.
The lowering of the resolution increases the effective size of the kernel by the same cumulative factor: $\prod^{\layer-1}_{i=0} \stridesize[i]$, where $\stridesize[i]$ represents the stride size of a layer $i$.
This allows us to compute the receptive field size of a layer $\layer{}$ using the following recursive formula:
\begin{equation}
   \rfsize[\layer] = r_{\layer - 1} + (\kernelsize[\layer{}] - 1) \prod^{\layer-1}_{i=0} \stridesize[i]
\end{equation}
We substitute $\Delta\rfsize[\layer{}] = (\kernelsize[\layer{}] - 1) \prod^{\layer-1}_{i=0} \stridesize[i]$. 
Since $r_0 = k_0$, we can rewrite the equation non-recursively:
\begin{equation}
   \rfsize[\layer] = k_0 + \sum^{\layer}_{i=1} \Delta\rfsize[\layer{}]
   = k_0 + \sum^{\layer}_{i=1} \Bigg[\kernelsize[i] \prod^{i-1}_{j=0} \stridesize[j]\Bigg]
\end{equation}
While this method allows us to determine the receptive field for sequences of layers like the VGG-network family, more recent architectures cannot be described adequately as a sequence of layers. For example, most architectures feature skip-connections, which effectively represent an alternative pathway through the network's structure.
In such scenarios, the receptive field is commonly computed from the path with the largest receptive field expansion \cite{araujo2019computing, floridaRF, inceptionv2, receptive_field_audio, sizematters, fpn}, ignoring the rest of the network's topology leading to the target layer.
We refer to this definition of the receptive field as $\rfsizemax[\layer{}]$. 
This value can be interpreted as the upper bound for the size of features extractable by the layer $\layer$.
However, with modern architectures being very deep, this value commonly exceeds the input resolution, making it not very useful for determining realistic limits of feature extraction within the architecture.
For example, the $\rfsizemax[\layer{}]$ of ResNet18's final layer is 435, which is almost double the $224 \times 224$ pixel input resolution. EfficientNetB7's final convolutional layer has a $\rfsizemax[\layer{}] = 3079$ while processing an $600 \times 600$ input. 
We find that the minimum receptive field size $\rfsizemin[\layer]$, a lower bound for the most local information, is more useful for diagnosing and resolving design inefficiencies in neural architectures \cite{goingdeeper}.
Since we assume the architecture to be non-recurrent, there is a finite set of sequences $S_{\layer{}}$ leading from the input to the target layer $\layer$. Computing all receptive field sizes results in \rfsize[\layer] being a set of values.
Since this set is always finite when the architecture is non-recurrent, we can trivially obtain the minimum receptive field size $\rfsizemin[\layer]$:
\begin{equation}
   \rfsizemin[\layer] = \min( \{ r_{\layer, s} \mid s \in S_{\layer{}} \}) 
\end{equation}
\begin{equation}
    = \min \Bigg(\Big\{k_0 + \sum^{s}_{i=1} \Bigg[\kernelsize[i] \prod^{i-1}_{j=0} \stridesize[j]\Bigg] \mid s \in S_{\layer{}} \Big\} \Bigg)
\end{equation}
This process is illustrated on a simple example in Fig. \ref{fig:catmp}, alongside the analog computation of the $\rfsizemax[\layer{}]$.

\subsection{Unproductive Layers and the Relationship of Input Resolution and Neural Architecture}
The minimum receptive field size $\rfsizemin[\layer{}]$ represents the lower bound for the locality of the features extractable by a layer $\layer$.
This implies that a layer should be unable to extract new features when it's input already has a receptive field size $\rfsizemin[\layer{}-1]$ as large or larger than the image: 
\begin{equation}
    \rfsizemin[\layer{}-1] \geq I_{res}
\end{equation}
where $(I_{res} \times I_{res})$ is the resolution of the input image.
While this analysis method is an oversimplification of the inference process, it was empirically demonstrated to have predictive power:
It has been shown that the linear separability of the latent representations does not improve in a CNN classifier when passed through a layer with $\rfsizemin[\layer{}-1] \geq I_{res}$ \cite{sizematters, goingdeeper, featurespace_saturation}.
We refer to such layers as unproductive layers, since they require computational resources at training and inference time, but do not produce measurable benefits in predictive quality, making them a localizable inefficiency within the architecture.
Removing unproductive layers has been observed to result in no reduction in predictive performance and an increase in computational and parameter efficiency \cite{goingdeeper, sizematters, featurespace_saturation}.

\subsection{Formalizing Receptive Field Analysis}
Previous works \cite{goingdeeper, sizematters, exploring} have explored unproductive layers primarily on low-resolution datasets like Cifar10. 
Since the tested CNN classifiers are usually optimized for ImageNet1k and other higher-resolution datasets, unproductive layers occur regularly and in great number, when trained on the datasets' low native resolution.
However, the resulting inefficiencies cannot be considered a flaw in the design of the investigated architectures, since the architectures are transplanted into a low-resolution setup they were not designed to handle. 
In this work, we are interested in looking for unproductive layers that occur in the setup the architectures were originally designed for.
To do this efficiently we propose a simple procedure, to diagnose inefficient architectures:
We define a fully utilized CNN architecture $A$ as an architecture which can fully utilize the receptive field expansion of every layer. This can be expressed using the following condition:
\begin{equation}
\forall \layer \in A: \rfsizemin[\layer] < \inputsize
\end{equation}
Since this definition is conditioned on the size of the input image \inputsize{}, which can be chosen freely, we can deduct a minimum input resolution for which this condition holds true:
\begin{equation}
I_{min} = \max(\{\rfsizemin[\layer]  \mid \layer{} \in A \})
\end{equation}
Essentially, a CNN-architecture is guaranteed to not have unproductive layers, when it is trained with an input size of $I_{min}$ or higher, since every layer can enrich its input with additional context. 
We choose $\rfsizemin[\layer]$ instead of $\rfsizemin[\layer - 1]$, to also exclude architectures with layers that partially exceed the boundaries of the image when expanding the receptive field. 
While these layers have been empirically shown to advance the quality of the prediction, their contributions seem to limited by not being able to fully utilize the receptive field expansion due to the minimum receptive field going beyond the images' borders \cite{sizematters, goingdeeper}.
Theoretically, this definition could also be expanded to an interval ($I_{min}, I_{max})$ instead of a lower bound, where the largest feasible input resolution $I_{max}$ describes the largest possible feature that can be extracted based on the maximum receptive field sizes $I_{max} = \max_l(r_{\layer, max})$. 
However, we neglect this factor in this work, since modern architectures usually feature large values for $I_{max}$ far beyond 2000 pixels, which exceeds commonly used input resolutions of classifiers by orders of magnitude. 
Thus, this work focuses exclusively on unproductive layers and on the input size lower bound $I_{min}$.

\section{Comparison with Shape Adaptor}

\section{Uncovering Design Inefficiencies in Popular Architectures}
\label{sec:flaws}

\begin{table}[ht!]
  \caption{\textbf{Our list of underutilized architectures} according to our analysis of the receptive field. Since $I_{min} \geq I_{res}$ for all networks, lowering $I_{min}$ through architectural changes will result in improving the predictive performance.}
      \label{tab:ineff}

\centering
\renewcommand{\arraystretch}{1.25}
\begin{tabular}{l|c|c
}
  \textbf{Architecture}
  &
  $I_{res}$ &
  $I_{min}$ 
   \\ \cline{1-3}  \noalign{\vskip\doublerulesep
         \vskip-\arrayrulewidth} \cline{1-3} 
  \textbf{MobileNet} \cite{mobilenet}&
   $224 \times224$ &
   $315\times315$ \\ \cline{1-3} 

  \textbf{MobileNetV3 large} \cite{mobilenetv3}&
  $224 \times224$ &
  $263\times263$  \\ \cline{1-3} 

  \textbf{MobileNetV3 small} \cite{mobilenetv3}&
  $224 \times224$ &
  $303\times303$ \\ \cline{1-3} 

  \textbf{EfficientNetB0} \cite{efficientnet}&
  $224 \times224$ &
   $299\times299$ \\ \cline{1-3} 

  \textbf{EfficientNetB1} \cite{efficientnet}&
  $240 \times240$ &
  $299\times299$  \\ \cline{1-3} 

  \textbf{EfficientNetB2} \cite{efficientnet}&
  $260 \times260$ &
  $299\times299$ \\ \cline{1-3} 
  
  \textbf{NASNet A (mobile)} \cite{nasnet}&
  $224 \times224$ &
  $327\times327$ \\ \cline{1-3} 
  
  \textbf{VGG19} \cite{vgg}&
  $224 \times224$ &
  $268\times268$ \\ \cline{1-3} 
  
  \textbf{MnasNet (0.5)} \cite{mnasnet}&
  $224 \times224$ &
  $283\times283$ \\ \cline{1-3} 
  
  \textbf{MnasNet (0.75)} \cite{mnasnet}&
  $224 \times224$ &
  $283\times283$ \\ \cline{1-3} 
  
  \textbf{MnasNet (1.0)} \cite{mnasnet}&
  $224 \times224$ &
  $283\times283$ \\ \cline{1-3} 
  
  \textbf{MnasNet (1.3)} \cite{mnasnet}&
  $224 \times224$ &
  $283\times283$ \\ \cline{1-3} 
  
  \textbf{ConvNeXt (T)} \cite{convnext}&
  $224 \times224$ &
  $224 \times224$ \\ \cline{1-3} 
  
  \textbf{ConvNeXt (S)} \cite{convnext}&
  $224 \times224$ &
  $224 \times224$  \\ \cline{1-3} 
  
  \textbf{ConvNeXt (M)} \cite{convnext}&
  $224 \times224$ &
  $224 \times224$ \\ \cline{1-3} 
  
  \textbf{ConvNeXt (L)} \cite{convnext}&
  $224 \times224$ &
  $224 \times224$ \\ \cline{1-3} 
  
  \textbf{ConvNeXt (XL)} \cite{convnext}&
  $224 \times224$ &
  $224 \times224$ 

\end{tabular}
\end{table}

\begin{table*}[htb!]
  \caption{\textbf{Our EfficientNet-variants trained on ImageNet1k} compared to the original models. Depressing $I_{min}$ has shown to be an effective strategy, both variants of lowering $I_{min}$ are outperforming the original models.}
  \label{tab:efficientNet}
    \centering
\definecolor{lg}{gray}{0.9}
\begin{tabular}{ l|l|c|c|c|c|c|c }
Architecture & Variant & Acc@1 & Acc@5 & FLOPs & Params & $I_{res}$ & $I_{min}$\\ \hline

EfficientNet B0 & original \cite{efficientnet} & 77.1\% & 93.3\% & 0.42B & 5.3M & 224 & 299\\
 & shorter \& wider (ours) & 78.3\% & 93.9\% & 0.72B & 5.5M & 224 & 171 \\ 
 & less downsampling  (ours) & \textbf{78.5\%} & \textbf{94.4\%} & 1.78B & 5.3M & 224 & 203 \\ \hline
 
EfficientNet B1 & original \cite{efficientnet} & 79.1\% & 94.4\% & 0.74B & 7.8M & 240 & 299\\
 & shorter \& wider (ours) & 79.6\% & 94.8\% & 1.43B & 7.7M & 240 & 235 \\
 & less downsampling  (ours) & \textbf{80.6\%} & \textbf{95.2\%} & 1.67B & 7.8M & 240 & 203   \\ \hline
 
EfficientNet B2 & original \cite{efficientnet} & 80.1\% & 94.9\% & 1.06B & 9.2M & 260 & 299\\
 & shorter \& wider (ours) & 80.7\% & 95.2\% & 2.27B & 9.6M & 260 & 235 \\
 & less downsampling  (ours) & \textbf{81.2\%} & \textbf{95.6\%} & 2.46B & 9.2M & 260 & 203  \\ \hline
 
EfficientNet B3 & original \cite{efficientnet} & 81.6\% & 95.7\% & 1.96B & 12M & 300 & 299\\
 & shorter \& wider (ours) & 81.8\% & 95.9\% & 3.72B & 12M & 300 & 235 \\
 & less downsampling (ours) & \textbf{82.8\%} & \textbf{96.0\%} & 4.41B & 12M & 300 & 203 \\ \hline

EfficientNet B4 & original \cite{efficientnet} & 82.9\% & 96.4\% & 4.66B & 19M & 380 & 299\\
 & shorter \& wider (ours) & \textbf{83.4\%} & 96.7\% & 9.27B & 19M & 380  & 235 \\
 & less downsampling  (ours) & 83.3\% & \textbf{96.7\%} & 11.2B & 19M & 380 & 203 \\ \hline
 
EfficientNet B5 & original \cite{efficientnet} & 83.6\% & 96.7\% & 10.8B & 30M & 456 & 299\\
 & shorter \& wider (ours) & 83.8\% & 96.7\% & 23.2B & 32M & 456 & 235 \\
 & less downsampling  (ours) & \textbf{83.8\%} & \textbf{96.8\%} & 25.8B & 30M & 456 & 203 \\ 

\end{tabular}
\end{table*}

We proceed to investigate popular architectures by computing their respective values for $I_{min}$.
In table \ref{tab:ineff} we provide selected results for architectures with $I_{min} \geq I_{res}$.
A full table of all tested architectures can be found in  table \ref{tab:inversRFA} in the supplementary material.
According to our previous definition, we consider the parameters in the architectures in table \ref{tab:ineff} underutilized.
These results suggest that gains in predictive performance should be possible for these architectures without increasing the number of parameters.

We hypothesize that this can be achieved by lowering $I_{min}$ such that $I_{min} < I_{res}$, without significantly changing the number of parameters in the architecture.
We propose two distinct procedures, illustrated in figure \ref{fig:overview}: 
First, reactivating underutilized layers by lowering $\rfsize[\layer{}, min]$, such that $\rfsize[\layer{}, min] < I_{min }$ for all previously underutilized layers $\layer$. 
This is easily achieved by reducing the downsampling in the architecture, decreasing the growth of the receptive field. 
Effectively, we change the stage-layout of the architecture's superstructure to slow the receptive field growth, resulting in full utilization of the architecture's parameters.
Alternatively, we remove the underutilized layers with $\rfsize[\layer, min] \geq I_{res}$ and distribute the parameters to preceding layers to keep the capacity of the model constant.
This can also be seen as a change in the network's superstructure, reducing the depth and increase the width of some stages without necessarily changing the design of the individual building blocks.

In both cases, the changes are very minimal and generally require the alteration of only a few lines of code. 
Such changes are also more economical regarding the computational and memory footprint than simply increasing $I_{res}$ to achieve the same difference between $I_{min}$ and $I_{res}$, since not the entire architecture's footprint is scaled quadratically.
To further make exploring the effects of the changes to $I_{min}$ as easy as possible, we provide code that automates the extraction of receptive field information directly from PyTorch-models and provide it in our supplementary material.

\subsection{Optimizing the EfficientNet Architectures}
\label{sec:effresolving}

\begin{table}
  \caption{\textbf{When training EfficientNet-models on ImageNet1k without scaling $I_{res}$}, our variants outperform the underutilized original consistently, demonstrating that scaling $I_{res}$ partially cannibalized the effect of depressing $I_{min}$ in table \ref{tab:efficientNet}.}
  \label{tab:lowResEffNet}
\definecolor{lg}{gray}{0.9}
\centering
\begin{tabular}{ l|l|l|l|l }

 & Var.&  Acc@1 & FLOPs & $I_{res}$\\ \hline
B0 & original \cite{efficientnet} & 77.1\% & 0.42B & 224 \\
 & shorter \& wider (ours) & 78.2\% & 0.84B & 224 \\
 & less downsampling  (ours) & \textbf{78.5\%} & 0.94B & 224 \\ \hline

B1 & original \cite{efficientnet} & 78.8\% & 0.62B & 224 \\
 & shorter \& wider (ours) & 79.4\% & 1.24B & 224 \\
 & less downsampling  (ours) & \textbf{80.3\%} & 1.45B & 224 \\ \hline

B2 & original \cite{efficientnet} & 79.6\% & 0.71B & 224 \\
 & shorter \& wider (ours) & 80.2\% & 1.56B & 224 \\
 & less downsampling (ours) & \textbf{80.9\%} & 1.71B & 224 \\ \hline

B3 & original \cite{efficientnet} & 80.5\% & 1.03B & 224 \\
 & shorter \& wider (ours) & 81.2\% & 2.03B & 224 \\
 & less downsampling (ours) & \textbf{81.8\%} & 2.40B & 224 \\ \hline

B4 & original \cite{efficientnet} & 81.2\% & 1.59B & 224 \\
 & shorter \& wider (ours) & 81.9\% & 3.17B & 224 \\
 & less downsampling (ours) & \textbf{82.5\%} & 3.83B & 224 \\ \hline
 
B5 & original \cite{efficientnet} & 81.7\% & 2.48B & 224 \\
 & shorter \& wider (ours) & 82.5\% & 5.42B & 224 \\
 & less downsampling  (ours) & \textbf{82.8\%} & 6.08B & 224 \\ \hline

B6 & original \cite{efficientnet} & 80.7\% & 3.52B & 224 \\
 & shorter \& wider (ours) & 82.3\% & 7.59B & 224 \\
 & less downsampling  (ours) & \textbf{82.9\%} & 8.68B & 224 \\ \hline

B7 & original \cite{efficientnet} & 81.8\% & 5.39B & 224 \\
 & shorter \& wider (ours) & 82.2\% & 11.4B & 224 \\
 & less downsampling  (ours) & \textbf{83.0\%} & 13.4B & 224 \\ 
\end{tabular}
\end{table}


From table \ref{tab:ineff}, we observe that EfficientNet B0, B1 and B2 are not fully utilized, since $I_{min} > I_{res}$.
EfficientNet models consist of 7 stages composed of a number of MBConvBlocks with the same number of filters and kernel-sizes. The first block within stages 2, 3, 4 and 6 are also downsampling the feature maps.
To investigate the effectiveness of lowering $I_{min}$, we test both strategies proposed in the previous section on EfficientNet.

\begin{table*}[htb!]
    \caption{\textbf{Classification accuracies of improved architectures on ImageNet1K}. In general, shrinking $I_{min}$ below the input resolution $I_{res}$ improves the predictive performance in all tested scenarios. This is true when comparing the performance to the original accuracy achieved by the respective authors and our reproduction using the same training setup as our improved variants.}
        \label{tab:improved}
        \centering
        \definecolor{lg}{gray}{0.9}
        \begin{tabular}{ l|l|c|c|c|c|c|c }
        Architecture & Variant & Acc@1 & Acc@5 & FLOPs & Params & $I_{res}$ & $I_{min}$\\ \hline
         
        VGG19  & original \cite{vgg} & 74.5\% & 92.0\% & 19.71B & 143.67M & 224 & 268 \\
         & reproduction  & 76.1\% & 92.9\% & 19.71B & 143.67M & 224 & 268 \\
         & reduced $I_{min}$ (ours)  & \textbf{76.3\%} & 93.0\% & 51.71B & 143.67M & 224 & 220 \\ \hline
         
        MobileNetV1 & original \cite{mobilenet} & 70.6\% & not provided & 0.59B & 4.23M & 224 & 315 \\
         & reproduction & 73.8\% & 91.5\% & 0.59B & 4.23M & 224 & 315 \\
         & reduced $I_{min}$ (ours)  & \textbf{74.6\%} & 92.0\% & 1.39B & 4.23M & 224 & 219 \\ \hline
         
        MobileNetV3 (small) & original \cite{mobilenetv3}\textbf{} & 67.4\% & not provided & 0.06B & 2.54M & 224 & 303 \\
         & reproduction & 67.2\% & 87.5\% & 0.06B & 2.54M & 224 & 303 \\
         & reduced $I_{min}$ (ours) & \textbf{70.6\%} & 89.4\% & 0.18B & 2.54M & 224 & 175\\ \hline
         
        MobileNetV3 (large) & original \cite{mobilenetv3}  & 75.2\% & not provided & 0.24B & 5.48M & 224 & 263 \\
         & pretrained (21K) \cite{imagenet21k} & 78.0\%  & not provided & 0.24B & 5.48M & 224 & 263 \\
         & reproduction & 77.1\% & 93.5\% & 0.24B & 5.48M & 224 & 263 \\
         & reduced $I_{min}$ (ours)  & \textbf{78.5\%} & 94.2\% & 0.38B & 5.48M & 224 & 199\\ \hline
         
        MNASNet(0.5) & original \cite{mnasnet} & 68.9\% & not provided & 0.12B & 2.21M & 224 & 283 \\
         & reproduction & 68.2\% & 87.9\% & 0.12B & 2.21M & 224 & 283 \\
         & reduced $I_{min}$ (ours)  & \textbf{69.8\%} & 88.8\% & 0.24B & 2.21M & 224 & 219 \\ \hline 
        
        MNASNet(0.75) & original \cite{mnasnet} & 73.3\% & not provided & 0.24B & 3.17M & 224 & 283\\
         & reproduction & 73.9\% & 91.1\% & 0.24B & 3.17M & 224 & 283\\
         & reduced $I_{min}$ (ours)  & \textbf{74.4\%} & 91.8\% & 0.47B & 3.17M & 224 & 219\\ \hline  
        
        MNASNet(1.0) & original \cite{mnasnet} & 75.2\% & not provided & 0.34B & 4.38M & 224 & 283\\
         & reproduction & 73.5\% & 91.1\% & 0.34B & 4.38M & 224 & 283\\
         & reduced $I_{min}$ (ours)  & \textbf{75.7\%} & 92.6\% & 0.73B & 4.38M & 224 & 219 \\ \hline
         
        MNASNet(1.3) & original \cite{mnasnet} & 77.2\% & not provided & 0.56B & 6.28M & 224 & 283\\
         &  reproduction & 76.5\% & 93.3\% & 0.56B & 6.28M & 224 & 283\\
         & reduced $I_{min}$ (ours)  & \textbf{77.4\%} & 93.6\% & 1.17B & 6.28M & 224 & 219  \\ \hline
         
        NASNet A (mobile) & original \cite{nasnet} & 74.0\% & 91.6\% & 0.59B  &  5.29M & 224 & 327 \\
         & reproduction & 75.1\% & 92.5\% & 0.59B  &  5.29M & 224 & 327 \\
         & reduced $I_{min}$ (ours)  & \textbf{78.0\%} & 93.9\% & 2.35B & 5.29M & 224 & 165 \\ 
    
    \end{tabular}
\end{table*}

We first test removing underutilized layers and making the architecture partially wider to account for the cut parameters:
We remove all building blocks featuring at least one layer with $\rfsizemin[\layer{}] \geq I_{res}$, which is the entire stage 7 and stage 6 except for the first building block.
To maintain the number of parameters of the original EfficientNetB0 architecture the number of filters in stage 5 are increased from 112 to 165 and in stage 6 from 192 to 320. We refer to this variant as ``shorter \& wider''. The second variant is reactivating underutilized layers by reducing downsampling, as described in the previous section. In this case, the downsampling is removed from the fourth stage of the model. This change is sufficient to lower the receptive field growth enough to make the architecture fully utilized. 
We refer to this variant as ``less downsampling''.
The described variants were made with three objectives in mind.
First, conduct minimal changes to the original architecture's structure. Second, depress the receptive field sizes enough such that $I_{min} < I_{res} = 224$ and finally for $I_{min}$ to be as close to $I_{res}$ as possible. 
The latter constraint is used to avoid some limitations of this analysis method.
The authors of \cite{convnext} have demonstrated that having very low receptive fields by training completely downsampling-free models has a negative effect on performance.
We therefore suspect that there is a soft lower-limit to beneficial effects of depressing the receptive field.
For this reason, we make changes that result in $I_{min}$ being as close to $I_{res}$ as possible while still being smaller.
To obtain modified variants of EfficientNetB1 and higher, we use the same scaling strategy as the original EfficientNet and apply it on our variants. 
Please note that the differently sized variants mostly share a value for $I_{min}$, which is due to skip connections in the building blocks providing pathways that do not expand the receptive field, thereby keeping $\rfsizemin[\layer{}]$ low.

From the results in table \ref{tab:efficientNet}, we observe that both modifications improved the top1 and top5 accuracy of the respective model.
In some cases, like EfficientNetB1 the reduced striding resulted in performance higher than the original B2-version of the model.
Interestingly, the computationally more expensive change yields higher performance gains.
Quintessentially, an increase in predicting performance and parameter-efficiency is achieved by increasing the computational footprint of the model.

EfficientNet models from B3 to B5 were fully utilized due to their higher $I_{res}$.
Our variants also outperform these models, but to a lesser degree, since the increasing $I_{res}$ is cannibalizing the effects of lowering $I_{min}$
This can be demonstrated further, by training all EfficientNet models by not scaling $I_{res}$ with the architecture. In this setup, all the original EfficientNet architectures are not fully utilized and are performing consistently worse than both improved versions (see table \ref{tab:lowResEffNet}). 
In both cases, the more computationally expensive changes that resolved underutilization yields higher accuracies, which is consistent with prior observations.
Finally, the depicted results outperform variants of EfficientNetB0 that use DynOPool \cite{rfpooling}, a dynamic pooling operator automatically adjusting the model's receptive field, which only achieved 72.8\% top1-accuracy for EfficientNetB0. We provide further comparisons with similar approaches like Shape Adaptor \cite{shapeadaptor} in the supplementary material.

\subsection{Optimizing other Popular Architectures}
\label{sec:restresolving}

To further demonstrate the effectiveness of our approach we test different modification procedures on the other architectures shown to be not fully utilized based on table \ref{tab:ineff}.
In this section, we primarily focus on changing the downsampling within an architecture, since these changes were more minimally invasive and yield higher predictive performance.
In an effort to further showcase the robustness of this procedure, we try in all cases to keep the alterations as minimal as possible to depress $I_{min}$, while also varying the approach from architecture to architecture.
For all tested models, we only ever started and completed training on a single modified variant and no hyperparameter optimization was conducted, making this refinement process completely trial-and-error free.
This is possible, since $I_{min}$ and $I_{res}$ are known before training, allowing to alter the architecture in the intended way without the need for training.
The changes made to each architecture are as follows:

\begin{table*}[htb!]
    \caption{\textbf{Classification accuracies of improved ConvNeXt on ImageNet1k}. Similar to other architectures, ConvNeXt improves substantially by lowering $I_{min}$.}
    \label{tab:convnext}

    \centering
    \definecolor{lg}{gray}{0.9}
    \begin{tabular}{ l|l|c|c|c|c|c|c }
        Architecture & Variant & Acc@1 & Acc@5 & FLOPs & Params & $I_{res}$ & $I_{min}$\\ \hline
        
        ConvNeXt T & original \cite{convnext} & 82.1\% & not provided & 4.46B & 28.58M & 224 & 224\\
         & reduced $I_{min}$ (ours)  & \textbf{83.1\%} & 96.4\% & 17.81B & 28.58M & 224 & 112 \\ \hline
        
        ConvNeXt S & original \cite{convnext} & 83.1\% & not provided & 8.70B & 50.22M & 224 & 224 \\
        & reduced $I_{min}$ (ours)  & \textbf{84.0\%}  & 96.6\% & 34.76B & 50.22M & 224 & 112 \\ \hline
        
        ConvNeXt B & original \cite{convnext} & 83.8\% & not provided & 15.38B & 88.59M & 224 & 224\\
         & reduced $I_{min}$ (ours)  & \textbf{84.3\%} & 96.9\% & 61.45B & 88.56M & 224 & 112 \\ 
         & original (finetuned) \cite{convnext} & 85.1\% & not provided & 45.19B & 88.59M & 384 & 224\\ 
         & reduced $I_{min}$ (ours) (finetuned)  & \textbf{85.5\%} & 97.5\% & 180.59B & 88.56M & 384 & 112 \\ 
          & original (21K finetuned) \cite{convnext} & 86.8\% & not provided & 45.19B & 88.59M & 384 & 224\\ 
         & reduced $I_{min}$ (21K finetuned) (ours)  & \textbf{87.1\%} & 98.2\% & 180.59B & 88.56M & 384 & 112 \\ 
        \hline

        ConvNeXt L & original \cite{convnext} & 84.3\% & not provided & 34.39B & 197.77M & 224 & 224 \\
         & reduced $I_{min}$ (ours) & \textbf{84.7\%} & 97.1\% & 137.49B & 197.76M & 224 & 112 \\
         & original (finetuned) \cite{convnext} & 85.5\% & not provided & 101.08B & 197.77M & 384 & 224\\ 
         & reduced $I_{min}$ (finetuned) (ours)  & \textbf{85.7\%} & 97.6\% & 404.06B & 197.76M & 384 & 112 \\ 

    \end{tabular}
\end{table*}

\begin{itemize}
    \setlength{\itemsep}{0pt plus 5pt}
    \item \textbf{VGG19} \hfill $\Delta I_{min}:= 269 \rightarrow 220$ \\  
    The first MaxPooling layer is placed after the fourth convolutional layer. \\ Afterwards, pooling is applied after very three convolutional layers. 
    \item \textbf{MobileNetV1}   \hfill
    $\Delta I_{min}:= 315 \rightarrow 219$ \\
    Downsampling occurs now in the layers 1, 3, 5 and 11 instead of 1, 3, 5, 7, 13. 
    \item \textbf{MobileNetV3 (Small)}  \hfill     $\Delta I_{min}:= 303 \rightarrow 175$ \\
    Changed the third downsampling layer's stride size from 2 to 1. 
     \item \textbf{MobileNetV3 (Large)} \hfill $\Delta I_{min}:= 263 \rightarrow 199$ \\
    Changed the last downsampling layer's stride size from 2 to 1. 
    
    \item \textbf{MNASNet (all variants)} \hfill     $\Delta I_{min}:= 283 \rightarrow 219$ \\
    Changed the last downsampling layer's stride size from 2 to 1. 
    \item \textbf{NASNet A (mobile)} \hfill $\Delta I_{min}:= 327 \rightarrow 165$ \\
    Changed stride size of the first stem layer from 2 to 1. 
     
    \item \textbf{ConvNeXt (all variants)} \hfill  $\Delta I_{min}:= 224 \rightarrow 112$ \\
    Changed the patching layer from $4 \times 4$ patches to $2 \times 2$ patches. 
   
\end{itemize}

We train all models on ImageNet1k using the standard configuration described in table \ref{tab:hyper} of the supplementary material. 
Additionally, to account for the influence on hyperparameters, we train the original architectures using the same setup and report the performance results alongside the results of the original authors and our improved models in table \ref{tab:improved}: For all tested models, our improved variant reliably outperforms the original model and our reproduction of the original model using our standard training setup.
Therefore, we see this as further confirmation that depressing $I_{min}$ below the input resolution to fully utilize the network is a robust analysis method for reliably optimizing architectures.
Finally, we would like to highlight the results of ConvNeXt in table \ref{tab:convnext}. 
Not only were we able to outperform ConvNeXt on all size variants using the standard ImageNet1K training, but also when fine-tuning ConvNeXt B and L on a higher resolution.
We were also able to outperform ConvNeXt B and L when pretraining on ImageNet21k (see supplementary material).
To our knowledge this is achieving a new state-of-the-art for CNN-models that did not use complex pretraining procedures \cite{coca}, distilling \cite{deit} or massive pretraining datasets like JFT300M \cite{vit, nfnet}.

\section{Summary and Discussion}
\label{sec:improved_models}
In conclusion, we have shown that the inductive bias of CNNs can be effectively used to guide the practitioner's design decisions regarding the network's superstructure.
We derive a simple but effective analysis method to guide the performance oriented refinement of CNN-superstructures.
Using this analysis method, we uncover design inefficiencies in various popular architectures, 
demonstrating the utility of finding and fixing issues with underutilized layers. Perhaps surprisingly, this analysis is even effective at identifying underutilized layers across a wide variety of the highest performing modern architectures.
%
By applying different strategies to resolve those inefficiencies we can improve the predictive performance of all tested models on the first attempt, achieving new state-of-the-art results in the process, without increasing the number of parameters in the model.
While we universally increased the parameter-efficiency of our models, this came at increased computational cost, indicating that an inherent trade-off between parameter and computational efficiency exists specific for each model class.
This is interesting from the perspective of scaling laws \cite{scalinglaws}, since we effectively demonstrate that parameters can be substituted for additional compute and do not necessarily need to be scaled together to be effective.
%


When practically applying these ideas to potential changes to a CNN architecture, decisions are influenced by current model inefficiencies, the superstructure of the architecture, the compute budget for training the model and the desired improvement in predictive performance.
In this context, the advantage of our approach is that choosing to spend additional computational resources for increased predictive performance can be made intentionally and reliably, not requiring expensive trial-and-error based comparative evaluation between model variants.

To make our receptive field analysis approach more useful and accessible we have created an implementation compatible with TensorFlow and PyTorch, which will be provided in our supplementary material and via PyPi.

\section{Acknowledgement}

This work was supported by a fellowship within the IFI program of the German Academic Exchange Service (DAAD).
We would also like to thank  Google’s TPU Research Cloud (TRC) for Cloud TPUs used in this research.
\bibliographystyle{unsrt}  
\bibliography{egbib.bib}

\clearpage

\appendix

\onecolumn

\section{Codebases for training models}
For training ConvNeXt we use the official repository published by the original authors: \\ \url{https://github.com/facebookresearch/ConvNeXt}. \\
The remaining models are trained using the training-script of the timm-library: \\ \url{https://github.com/rwightman/pytorch-image-models.git}.

\section{Training Setups}
In this section, we briefly discuss the training setup used for all trainings throughout this work.
The first configuration in table \ref{tab:hyper} is strongly inspired by the A2 hyperparameter-configuration by \cite{strikesback}.
We use this configuration, since it reflects a state-of-the-art setup for CNN-classifiers in computer vision \cite{nfnet, convnext, efficientnet}. 
It is also known to generalize well over a wide set of models, achieving very competitive performance on old and new architectures \cite{strikesback}.
For this reason, we see this setup as a good common testing ground to evaluate the effects of our architectural changes.

Since ConvNeXt is the only model performing substantially worse on this configuration, we use its original setup and codebase \cite{convnext} for training, pre-training and fine-tuning. 
This also allows us to very directly compare our improved version with the original models.

However, when pretraining on ImageNet21K, there is one key-derivation:
The original authors use the Fall11 release of the unprocessed dataset, which is no longer available. We use the Winter21-version of ImageNet21K-P, which was the only version of this dataset available to us.
Based on the ablative studies of \cite{imagenet21k} on the different releases of ImageNet21K this change likely reduced the performance of our pretrained models slightly (see supplementary material of \cite{imagenet21k}).
However, since our models still outperform the original architectures we argue that the results still support the claims made in this work.

\begin{table}[htb!]
\caption{Hyperparameter settings used throughout this paper. The standard setup did not produce a competitive result for ConvNeXt-models, which is the reason why it is trained on its original setup and codebase published by the authors \cite{convnext}.}
\centering
\label{tab:hyper}
\begin{tabular}{ l | c c c c}
  training config & Standard Setup & ConvNeXt & ConvNeXt (finetuning 1K) & ConvNeXt (finetuning 21K) \\ \hline
  weight init &  $\mathcal{U}(-\sqrt{k}, \sqrt{k})$ & trunc. normal (0.2) & finetuning & finetuning\\
  optimizer & Lamb & AdamW & AdamW & AdamW\\
  learning rate & 5e-3  & 4e-3 & 5e-5 & 5e-5\\
  lr. decay & cosine & cosine & cosine & cosine\\
  weight decay & 0.02  & 0.02 & 1e-8 & 1e-8  \\
  opt. momentum & - & $\beta_1, \beta_2 = 0.9, 0.999$ & $\beta_1, \beta_2 = 0.9, 0.999$ & $\beta_1, \beta_2 = 0.9, 0.999$ \\
  batch size & 2048 & 4096 & 512 & 512 \\
  epochs & 300 & 300 & 30 & 30\\
  warmup epochs & 5 & 20 & 0 & 0\\
  warmup sched.  & linear & linear & N/A & N/A \\
  randaug \cite{random_augment} & None & (9, 0.5) & (9, 0.5) & (9, 0.5) \\
  mixup \cite{mixup} & 0.1 & 0.8 & None & None \\
  cutmix \cite{cutmix} & 1.0 & 1.0 & None & None\\
  rand. erasing \cite{randerase} & 0.0 & 0.25 & 0.25 & 0.25  \\
  color jitter & 0.4 & None & None & None \\
  crop pct. & 0.95 & 0.95 & 0.95 & 0.95 \\
  auto aug. \cite{autoaugment} & False & True & True & True \\
  smoothing \cite{inceptionv2} & 0.0 & 0.1 & 0.1 & 0.1 \\
  stoc. depth \cite{stochastic_depth} & 0.2 & 0.1/0.4/0.5/0.5  & 0.8/0.95 & 0.2 \\
  Loss & CE & BCE & BCE & BCE\\
\end{tabular}

\end{table}

\clearpage

\section{Computing $I_{min}$ and $I_{res}$ for various popular architectures.}
In table \ref{tab:ineff}, in the main part of this work, we present the respective $I_{min}$ for various popular architectures with $I_{min} \geq I_{res}$. In table \ref{tab:inversRFA} we present the results for all architectures tested during this work. 

\begin{table*}[htb!]
  \caption{Minimum feasible input resolution $I_{min}$ and the design resolution $I_{res}$ of popular CNN architectures. Architectures with $I_{min} \geq I_{res}$ are not fully utilizing all layers and are marked red in this table.}
  \label{tab:inversRFA}
\resizebox{\textwidth}{!}{
\renewcommand{\arraystretch}{1.25}
\begin{tabular}{
>{\raggedright\arraybackslash}p{0.25\linewidth}|
>{\centering\arraybackslash}p{0.15\linewidth}|
>{\centering\arraybackslash}p{0.15\linewidth}|
>{\centering\arraybackslash}p{0.001\linewidth}|
>{\raggedright\arraybackslash}p{0.25\linewidth}|
>{\centering\arraybackslash}p{0.15\linewidth}|
>{\centering\arraybackslash}p{0.15\linewidth}|
}
  \textbf{architecture}
  &
  $I_{res}$ &
  $I_{min}$ &
  &
  \textbf{architecture}
  &
  $I_{res}$ &
  $I_{min}$   \\ \cline{1-3} \cline{5-7} \noalign{\vskip\doublerulesep
         \vskip-\arrayrulewidth} \cline{1-3} \cline{5-7}
  \textbf{DenseNet121} \cite{densenet} &
  $224 \times224$ &
  $103\times103$ &
  &
  \cellcolor{red!25} \textbf{MobileNet} \cite{mobilenet}&
  \cellcolor{red!25} $224 \times224$ &
  \cellcolor{red!25} $315\times315$ \\ \cline{1-3} \cline{5-7}

  \textbf{DenseNet161} \cite{densenet}&
  $224 \times224$ &
  $103\times103$ &
  &
  \textbf{MobileNetV2} \cite{mobilenetv2}&
  $224 \times224$ &
  $163\times163$  \\ \cline{1-3} \cline{5-7}

  \textbf{DenseNet169} \cite{densenet}&
  $224 \times224$ &
  $103\times103$ &
  &
  \cellcolor{red!25} \textbf{MobileNetV3 large} \cite{mobilenetv3}&
  \cellcolor{red!25} $224 \times224$ &
  \cellcolor{red!25} $263\times263$  \\ \cline{1-3} \cline{5-7}

  \textbf{DenseNet201} \cite{densenet}&
  $224 \times224$ &
  $103\times103$ &
  &
  \cellcolor{red!25} \textbf{MobileNetV3 small} \cite{mobilenetv3}&
  \cellcolor{red!25} $224 \times224$ &
  \cellcolor{red!25} $303\times303$ \\ \cline{1-3} \cline{5-7}

  \cellcolor{red!25} \textbf{EfficientNetB0} \cite{efficientnet}&
  \cellcolor{red!25} $224 \times224$ &
   \cellcolor{red!25} $299\times299$ &
  &
  \textbf{Xception} \cite{xception}&
  $299 \times299$ &
  $135\times135$ \\ \cline{1-3} \cline{5-7}

  \cellcolor{red!25} \textbf{EfficientNetB1} \cite{efficientnet}&
  \cellcolor{red!25} $240 \times240$ &
  \cellcolor{red!25} $299\times299$ &
  &
  \textbf{NASNet A (large)} \cite{nasnet}&
  $331 \times331$ &
  $327\times327$ \\ \cline{1-3} \cline{5-7}

  \cellcolor{red!25} \textbf{EfficientNetB2} \cite{efficientnet}&
  \cellcolor{red!25} $260 \times260$ &
  \cellcolor{red!25} $299\times299$ &
  &
  \cellcolor{red!25} \textbf{NASNet A (mobile)} \cite{nasnet}&
  \cellcolor{red!25} $224 \times224$ &
  \cellcolor{red!25} $327\times327$ \\ \cline{1-3} \cline{5-7}

  \textbf{EfficientNetB3} \cite{efficientnet}&
  $300 \times300$ &
  $299\times299$ &
  &
  \textbf{ResNet18} \cite{resnet}&
  $224 \times224$ &
  $139\times139$ \\ \cline{1-3} \cline{5-7}

  \textbf{EfficientNetB4} \cite{efficientnet}&
  $380 \times380$ &
  $299\times299$ &
  &
  \textbf{ResNet34} \cite{resnet}&
  $224 \times224$ &
  $139\times139$ \\ \cline{1-3} \cline{5-7}

  \textbf{EfficientNetB5} \cite{efficientnet}&
  $456 \times456$ &
  $299\times299$ &
  &
  \textbf{ResNet50\textsuperscript{$\dagger$} \cite{resnet}} &
  $224 \times224$ &
  $96\times96$  \\ \cline{1-3} \cline{5-7}

  \textbf{EfficientNetB6} \cite{efficientnet}&
  $528 \times528$ &
  $299\times299$ &
  &
  \textbf{ResNet101\textsuperscript{$\dagger$} \cite{resnet}} &
  $224 \times224$ &
  $96\times96$ \\ \cline{1-3} \cline{5-7}

  \textbf{EfficientNetB7} \cite{efficientnet}&
  $600 \times600$ &
  $299\times299$ &
  &
  \textbf{ResNet152\textsuperscript{$\dagger$} \cite{resnet}} &
  $224 \times224$ &
  $96\times96$ \\ \cline{1-3} \cline{5-7}
  
  \textbf{VGG11} \cite{vgg}&
  $224 \times224$ &
  $150\times150$ &
  &

  \textbf{GoogLeNet} \cite{googlenet}&
  $224 \times224$ &
  $123\times123$ \\ \cline{1-3} \cline{5-7}


  \textbf{VGG13} \cite{vgg}&
  $224 \times224$ &
  $156\times156$ &
  &

  \textbf{SqueezeNet 1.0} \cite{squeezenet}&
  $224 \times224$ &
  $67\times67$ \\ \cline{1-3} \cline{5-7}

  \textbf{VGG16} \cite{vgg}&
  $224 \times224$ &
  $212\times212$ &
  &
  \textbf{SqueezeNet 1.1} \cite{squeezenet}&
  $227 \times227$ &
  $63\times63$ \\ \cline{1-3} \cline{5-7}

  \cellcolor{red!25} \textbf{VGG19} \cite{vgg}&
  \cellcolor{red!25} $224 \times224$ &
  \cellcolor{red!25} $268\times268$ &
  &
  \textbf{InceptionResNetV2} \cite{inceptionv4}&
  $299 \times299$ &
  $143\times143$ \\ \cline{1-3} \cline{5-7}

  \cellcolor{red!25} \textbf{MnasNet (0.5)} \cite{mnasnet}&
  \cellcolor{red!25} $224 \times224$ &
  \cellcolor{red!25} $283\times283$ &
  &
  
  \cellcolor{red!25} \textbf{ConvNeXt (T)} \cite{convnext}&
  \cellcolor{red!25} $224 \times224$ &
  \cellcolor{red!25} $224 \times224$ \\ \cline{1-3} \cline{5-7}
  
  \cellcolor{red!25} \textbf{MnasNet (0.75)} \cite{mnasnet}&
  \cellcolor{red!25} $224 \times224$ &
  \cellcolor{red!25} $283\times283$ &
  &
  \cellcolor{red!25} \textbf{ConvNeXt (S)} \cite{convnext}&
  \cellcolor{red!25} $224 \times224$ &
  \cellcolor{red!25} $224 \times224$  \\ \cline{1-3} \cline{5-7}
  
  \cellcolor{red!25} \textbf{MnasNet (1.0)} \cite{mnasnet}&
  \cellcolor{red!25} $224 \times224$ &
  \cellcolor{red!25} $283\times283$ &
  &
  \cellcolor{red!25} \textbf{ConvNeXt (M)} \cite{convnext}&
  \cellcolor{red!25} $224 \times224$ &
  \cellcolor{red!25} $224 \times224$ \\ \cline{1-3} \cline{5-7}
  
  \cellcolor{red!25} \textbf{MnasNet (1.3)} \cite{mnasnet}&
  \cellcolor{red!25} $224 \times224$ &
  \cellcolor{red!25} $283\times283$ &
  &
  \cellcolor{red!25} \textbf{ConvNeXt (L)} \cite{convnext}&
  \cellcolor{red!25} $224 \times224$ &
  \cellcolor{red!25} $224 \times224$ \\ \cline{1-3} \cline{5-7}
  
\textbf{InceptionV3} \cite{inceptionv3}&
  $299 \times299$ &
  $239\times239$ &
  &
  \cellcolor{red!25} \textbf{ConvNeXt (XL)} \cite{convnext}&
  \cellcolor{red!25} $224 \times224$ &
  \cellcolor{red!25} $224 \times224$ \\ \cline{1-3} \cline{5-7}

\end{tabular}}
\end{table*}

\clearpage

\section{Exploring Finetuning Procedures Using our Improved ConvNeXt B}
In this section, we briefly explore the effects of different finetuning procedures on our improved version of ConvNeXt, to demonstrate the robustness of our approach.
The depicted examples of pretraining and finetuning are using the same configurations and procedures described by the original authors \cite{convnext}.
We compare our model with the results of the original authors in table \ref{tab:finetune}.
From the results we observe that our model is able to outperform ConvNeXt B in all tested scenarios, which involves training from scratch as well as pretraining from ImageNet1K and 21K.
Please note that we do not test ConvNeXt L, which we briefly mentioned in the main text of this work, this is an unfortunate typo.

\begin{table*}[htb]
    \caption{\textbf{Classification accuracies of improved ConvNeXt B on ImageNet1K}. Improvements from lowering $I_{min}$ can be observed using different pretraining and finetuning procedures.}
    \label{tab:finetune}
    \centering
\definecolor{lg}{gray}{0.9}
\begin{tabular}{ l|l|c|c|c|c|c|c }
Architecture & Variant & Acc@1 & Acc@5 & FLOPs & Params & $I_{res}$ & $I_{min}$\\ \hline

ConvNeXt B & original \cite{convnext} & 83.8\% & not provided & 15.38B & 88.59M & 224 & 224\\
(training from scratch) & reduced $I_{min}$ (ours)  & \textbf{84.3\%} & 96.9\% & 61.45B & 88.56M & 224 & 112 \\ 
 \hline
 
ConvNeXt B & original \cite{convnext} & 85.1\% & not provided & 45.19B & 88.59M & 384 & 224 \\ 
(pretraining on ImageNet1K) & reduced $I_{min}$ (ours) & \textbf{85.5\%} & 97.5\% & 180.59B & 88.56M & 384 & 112 \\ 
\hline

 ConvNeXt B & original \cite{convnext} & 85.8\% & not provided & 15.38B & 88.59M & 224 & 224\\ 
(pretraining on ImageNet21K-P) & reduced $I_{min}$ (ours)& \textbf{86.0\%} & 97.9\% &  61.45B & 88.59M & 224 & 112 \\ 
\hline

ConvNeXt B & original \cite{convnext} & 86.8\% & not provided & 45.19B & 88.59M & 384 & 224\\ 
(pretraining on ImageNet21K-P) & reduced $I_{min}$ (ours)& \textbf{87.1\%} & 98.2\% & 180.59B & 88.56M & 384 & 112 \\  

\end{tabular}
\end{table*}

\section{Comparison to DynOPool and Shape Adaptor}
\label{sec:shapeadaptor}

\begin{table*}[htb]
    \caption{Comparison of DynOPool and Shape Adaptor with our approach.}
    \label{tab:rfpoolcomp}
    \centering
\definecolor{lg}{gray}{0.9}
\begin{tabular}{ l|l|c|c }
Model & Method & Cifar100 & Cifar10 \\ \hline

ResNet50 & DynOPool \cite{rfpooling} & \textbf{80.6\%} & not provided  \\
 & Shape Adaptor \cite{shapeadaptor} & 80.3\% & 95.5\%  \\
  & Human \cite{shapeadaptor} & 78.5\% & 95.5\%  \\
 & reduced $I_{min}$ (ours)  & 78.8\% & \textbf{95.9\%}  \\ \hline
 
 VGG16 & DynOPool \cite{rfpooling} & \textbf{79.8\%} & not provided  \\
 & Shape Adaptor & 79.2\% & 95.4\% \\
 & Human \cite{shapeadaptor} & 75.4\% & 94.1\%  \\
 & reduced $I_{min}$ (ours)  & 77.9\% & \textbf{95.7}\%  \\ \hline
 
 MobileNetV2 & DynOPool \cite{rfpooling} & 76.2\% & not provided  \\
 & Shape Adaptor \cite{shapeadaptor} & 75.7\% &93.9\%   \\
 & Human \cite{shapeadaptor} & 73.8\% & 93.7\%  \\
 & reduced $I_{min}$ (ours)  & \textbf{78.7\%} &  \textbf{95.3\%}  \\ 

\end{tabular}
\end{table*}

DynOPool  \cite{rfpooling} as well as Shape Adaptor \cite{shapeadaptor} directly adapt the receptive field by altering the downsampling procedures during training without changing the number of parameters.
In this section, we briefly compare the effectiveness of our guided manual approach to these automated procedures.

For this comparison, we use Cifar10 and Cifar100, training at the datasets' native resolution of $32 \times 32$ pixels. 
There are two reasons for this: First DynOPool as well as Shape Adaptor focus on low-resolution datasets in their experiments. Second, all three architectures used by the authors of both works (ResNet50, MobileNetV2 and VGG16) have $I_{min} < I_{res}$ for their respective ImageNet1K setups (see table \ref{tab:inversRFA}), which makes them fully utilized according to our definition.
However, $I_{res} = 32 \leq I_{min}$  for Cifar10 and Cifar100, making all three architectures very inefficient in this setup, allowing us to optimize these architectures.
For training, we reuse the same setup as for our other experiments, reducing the batch size to 128 images. 
Like in our previous experiments, we only implement and train a single model variant for each architecture.
We train improved variants with $I_{min} < I_{res}$ for MobileNetV2, ResNet50 and VGG16 and compare the results to DynOPool and Shape Adaptor variants in table \ref{tab:rfpoolcomp}. 

From the results we observe that our approach generally provides an improvement over the human baseline and also improves upon DynOPool and Shape Adaptor.
However, in two cases, Resnet50 and VGG16 on Cifar100, it was impossible to provide similar predictive performance, even when trying to match $I_{min}$ of our variant with $I_{min}$ of the Shape Adaptor variant of the same model, this performance could not be achieved.
We attribute this to the fact that the automated approaches use interpolation instead of downsampling, allowing the model to more gradually increase the receptive field growth.
This could be interesting for future work, since interpolation as a downsampling layer is not commonly used outside the investigated approaches.
Interestingly, all models produced by Shape Adaptor that we investigated in greater detail provided architectures with $I_{min} < I_{res}$, which we see as a confirmation of our approach.
The exact modification for refining the architecture to be fully utilized in Cifar10 and Cifar100 can be found below: \\

\begin{itemize}

    \setlength{\itemsep}{0pt plus 5pt}
    \item \textbf{ResNet50} \hfill     $\Delta I_{min}:= 96 \rightarrow 6$ \\
     The stem is replaced with a $3 \times 3$ convolution, mimicking the original Cifar-ResNet by \cite{resnet}.
    The only downsampling block left is the first bottleneck block with 1024 output filters / 256 filters in the $3 \times 3$ convolution.
    
    \item \textbf{VGG16} \hfill $\Delta I_{min}:= 212 \rightarrow 30$ \\
    Removed all pooling layers. Placed a single MaxPooling-Layer before the 3 final Conv-Layers of the feature extractor.
     
    \item \textbf{MobileNetV2} \hfill  $\Delta I_{min}:= 163 \rightarrow 29$ \\
    Removed downsampling from all layers except the one closest to the output. 
   
\end{itemize}

\clearpage

\section{Receptive Field Analysis Toolbox}
We provide RFA-Toolbox \url{https://anonymous.4open.science/r/receptive_field_analysis_toolbox-1CDD/README.md}, a simple package to conduct receptive field analysis.
Since the implementation has been a substantial amount of work and required (among other things) the reverse engineering of undocumented functions of the PyTorch-JIT-compiler, we provide some more detailed overview over the package:

\subsection{Overview}
\label{sec:overview}

\begin{figure*}[htb!]
    \centering
    \includegraphics[scale=0.7]{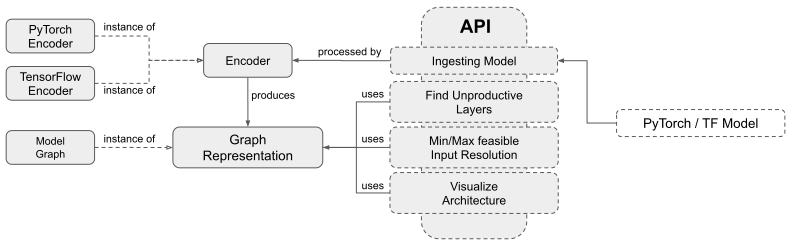}
    \caption{Software Architecture of RFA-Toolbox.}
    \label{fig:toolbox_arc}
\end{figure*}

In this section, we provide an overview on the inner workings of our tooling for extracting receptive field information from neural architectures.
Before we discuss the individual functionalities in greater detail, we provide a more general overview over RFA-Toolbox to put the components into context.
The structure of the Python-package is depicted in figure \ref{fig:toolbox_arc}.
The key component of RFA-Toolbox is a proprietary model-graph, which contains information critical for computing the receptive field.
All other functionalities of RFA-Toolbox operate directly on this graph structure instead of the original model.
Any model can be directly implemented in a declarative style using the proprietary graph of RFA-Toolbox.
However, we primarily see this as a fallback-option, since RFA-Toolbox also allows ingesting architectures from TensorFlow and PyTorch directly from the API, creating a proprietary graph in the process.
Obtaining a graph from a model is handled by a set of encoders, which are specialized to handle one specific framework each. 
This makes RFA-Toolbox independent of the framework of the model. It also allows us to add support for additional frameworks in the future without needing to implement functionalities of the API for multiple frameworks separately.
Besides the aforementioned processing of models into the proprietary graphs, all other exposed functions of the API are operating on a provided graph-instance to extract a specific piece of information like obtaining all nodes of unproductive and underutilized layers.
In summary, the API of the RFA-Toolbox currently provides four basic functionalities:

\begin{enumerate}
    \item Ingesting a model from a known framework like PyTorch and TensorFlow \cite{pytorch, tensorflow} and extract relevant information into a proprietary graph representation (see section \ref{sec:ingest})
    \item Finding unproductive and underutilized layers in a given architecture based on our analysis method (see section \ref{sec:example})
    \item Compute $I_{min}$ (see section \ref{sec:example})
    \item Visualizing the architecture and highlighting unproductive and underutilized layers (see section \ref{sec:example})
\end{enumerate}

\subsection{Ingesting Models into RFA-Toolbox}
\label{sec:ingest}
One of the key reasons why RFA-Toolbox is easy to use is that models from common frameworks like PyTorch and TensorFlow can directly be processed.
We refer to this process of extracting information about the architecture into our proprietary graph-structure as \textit{ingesting} a model.
We will first describe the proprietary graph-structure in greater detail, which is used for all internal processing.
This is followed by a description of the different encoders' implementations, which handle the process of ingestion for TensorFlow and PyTorch respectively.

\subsection{Proprietary Graph Structure}
The proprietary graph structure is not a fully functioning framework for implementing neural networks. Only some critical peaces of information are extracted from the original model. 
This involves the overall topology of the model as well as information that is either relevant for the receptive field computation like kernel and stride sizes or commonly depicted in visualizations like the name of the layer, the number of units etc.
However, the amount of information may be extended in the future to accommodate decoding a (potentially optimized) architecture back to a functioning framework.
One other important functionality of the graph is the receptive field computation. All necessary computations to obtain all receptive field sizes from all sequences leading to a node are done automatically during the instantiation of each node, based on the receptive-field information held by the node's direct predecessors. 
The graph is constructed bottom-up from the input to one or more outputs. 
Models with more than a single input are currently not supported. 
Edges are implicitly defined by lists containing pointers to direct predecessor and successor nodes held by each individual node.
The nodes also contain functionality that allows them to decide whether they are unproductive and underutilized layers.

\subsubsection{Keras / Tensorflow}
The encoder for TensorFlow is compatible exclusively with \code{keras.Model}, which is currently the default way of implementing models in TensorFlow.
RFA-Toolbox obtains the structure and meta-information from the \code{keras.Model} by decoding its json-String representation. 
This decoding process is based on a set of handlers that are called for each node in a specific sequence.
\label{sec:tensorflow}

\begin{figure*}[htb!]
    \centering
    \includegraphics[width=0.5\columnwidth]{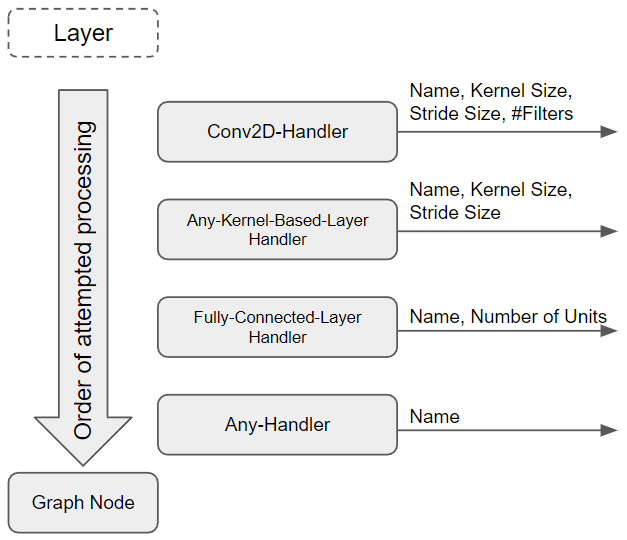}
    \caption{The information extraction process in the Keras and PyTorch-Encoders.}
    \label{fig:handlers}
\end{figure*}

Each handler can process a defined set of nodes in the compute-graph, and every node will be processed by exactly one handler.
The handlers will attempt to process the node in the order of their specificity. 
More specific handlers can only process a narrow subset of layers, for example only convolutional layers, but can extract more metadata from the respective layers. 
This process is illustrated in figure \ref{fig:handlers}.
The final handler is capable of processing any node, but will not extract any information from that node except for the name.
Layers that are processed like this are treated like convolutional layers with a kernel and stride size of 1 on all axes.
This is done to make these layers neutral for the computation of the receptive field and the future growth of the receptive field. 
By studying various implementations of models in TensorFlow, we find that this process can handle all pre-implemented classifier networks in the TensorFlow library and compute their receptive field sizes correctly.
The aforementioned list of handlers is mutable and can be expanded and modified manually by the user if required.
This could be necessary if novel custom layers are not processed correctly or require certain additional information to be extracted.

\subsubsection{PyTorch}
\label{sec:pytorch}
Different from TensorFlow, PyTorch operates on a dynamically generated compute graph, which cannot be trivially extracted from the model itself.
Furthermore, \code{torch.nn.Module}-instances, the basic building blocks of neural networks in PyTorch, may contain additional modules and complex processing logic.
Thus, to obtain the structure of the compute graph, we need to partially reverse engineer the JIT-compiler of PyTorch.
The JIT-compiler of PyTorch can trace the path of a tensor of data through the neural network, thereby constructing a graph of all layers and function calls that are involved in the processing of this tracing-tensor.
By using the string-representation of this JIT-compiled graph, we can reconstruct the order of function-calls and modules that processed the data. In essence, the graph allows us to reconstruct the neural architecture's topology.
From the naming of the nodes in the compute graph, we can map the nodes of the compute graph to the \code{torch.nn.Module}-instances in the network.
It is also possible to obtain function-calls and their parameters from the graph. 
This is crucial when functional-versions of layers that can affect the receptive field analysis, like pooling or convolutional layers, are used instead of their \code{torch.nn.Module}-based counterparts.
However, one minor draw-back of this method is that many nodes in the compute-graph are present for purely technical reasons, making the resulting graph very hard to read and more complex than necessary.
To enhance the readability of the graph, multiple post-processing steps remove such nodes from the graph.
These post-processing steps are not removing any computation that would affect the result of the receptive field analysis, so the results are consistent with the same model implemented in TensorFlow.
The extraction of meta-information, while technically more complex, is utilizing the same handler-pattern described for the processing of TensorFlow-models.
Additionally, due to the nesting of modules, RFA-toolbox provides a simple interface as well to allow the easy modification of handlers for PyTorch models.
This allows the user to view PyTorch-models in different levels of abstraction, which may be useful when visualizing very complex architectures.

However, the reliance of this process on the JIT-compiler also puts limitations on the functionality of the encoder. Since the graph-structure is dependent on the trace of a data-tensor being propagated through the network, structural changes that occur based on the state of the model are only reflecting the state of the model at the time the tracing took place.
This means that only the computations and modules in the control-flow are recognized that are executed by the traced data. Loops are entirely incompatible with this kind of extraction process.
However, while this seems like a severe limitation, only a single model-family (DenseNet by \cite{densenet}) from the torchvision-library is incompatible with RFA-Toolbox due to these limitations.

\subsection{Predicting and Visualizing Unproductive and Underutilized Layers}
\label{sec:example}
\begin{figure*}[ht!]
	\centering
	    \includegraphics[scale= 0.5]{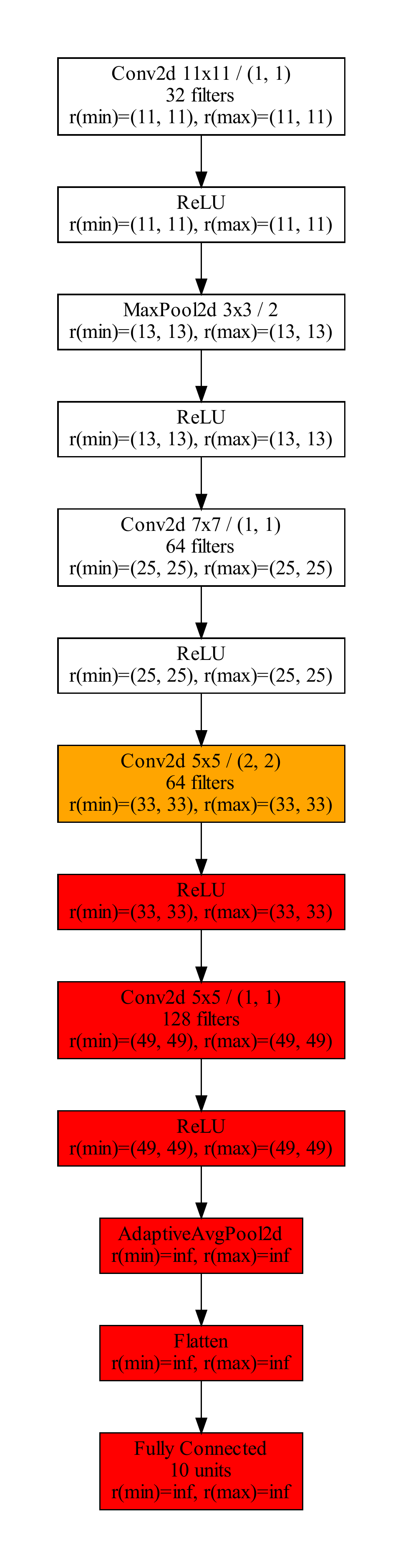}
	\vspace{-0.3cm}
	\caption{Visualization of a simple example architecture implemented in PyTorch. Unproductive layers for an input resolution of $32 \times 32$ pixels are colored red, underutilized layers are colored orange.}
	\label{fig:vgg_example}
\end{figure*}

Based on the graph extracted from the model, unproductive layers can be predicted using receptive field analysis.
Most importantly $I_{min}$ can be computed for any non-recurrent graph.
From the API, it is also possible to obtain lists of nodes in the graph with $I_{res} <= r_{\layer-1, min}$. Based on the layer's names, these nodes can be traced back to the layers, modules and function-calls in the original model.
This allows the user to embed RFA-Toolbox in automated systems like Neural Architecture Search to filter or optimize architectures with predicted inefficiencies.
Alternatively, RFA-Toolbox also supports a manual approach by providing functionality to visualize architectures and highlight unproductive layers (see figure \ref{fig:vgg_example}).
Additional to unproductive layers, we also provide an additional layer-state in this work for visualization purposes. Underutilized layers are layers with $I_{res} \leq r_{\layer, min}$. This means that they expand the receptive field past the boundaries of the image. Depending on the exact values $r_{\layer-1}$ and $I_{res}$ and $r_{\layer}$ this can mean that this layer is not utilizing its full potential. For example, when $r_{l-1}$ is very close to the input resolution $I_{res}$, the layer cannot integrate much novel context by expanding the receptive field.

\end{document}

%% file: latex/fi-opt.tex

\newcommand{\tikzrfaimage}[1][]{
    \fill[background] (-\rectsize,-\rectsize) rectangle (\rectsize,\rectsize);
    \draw[imagearea] (-\imgsize,-\imgsize) coordinate (Img1)
      rectangle (\imgsize,\imgsize) coordinate (Img2);
    \pic[#1] {owl};
    \draw[C,line width=1.6pt] (-\Csize,-\Csize) coordinate (C1)
      rectangle (\Csize,\Csize) coordinate (C2);
    \draw[B,line width=1.2pt] (-\Bsize,-\Bsize) coordinate (B1)
      rectangle (\Bsize,\Bsize) coordinate (B2);
    \draw[A,line width=1.0pt] (-\Asize,-\Asize) coordinate (A1)
      rectangle (\Asize,\Asize) coordinate (A2);
}
\newcommand{\tikzrfalayerA}[1]{
    \coordinate (left#1) at (0,0);
    \coordinate (right#1) at (.2,0);
    \draw[help lines,step=.2,shift={(0,.1)}] (0,-2.4) grid (0.2,2.2);
    \draw[A,line width=.6pt]
      (A1-|left#1) rectangle (A2-|right#1);
    \draw[B,line width=.6pt]
      (B1-|left#1) rectangle (B2-|right#1);
    \draw[C,line width=.6pt]
      (C1-|left#1) rectangle (C2-|right#1);
     
}
\newcommand{\tikzrfalayerB}[1]{
    \coordinate (left#1) at (0,0);
    \coordinate (right#1) at (.2,0);
    \draw[help lines,step=.2,shift={(0,.1)}] (0,-1.2) grid (0.2,1.0);
    \draw[A,line width=.6pt]
      (0,-.9) coordinate (A1L1) rectangle (.2,.9) coordinate (A2L1);
    \draw[B,line width=.6pt]
      (0,-.5) coordinate (B1L1) rectangle (.2,.5) coordinate (B2L1);
    \draw[C,line width=.6pt]
      (0,-.1) coordinate (C1L1) rectangle (.2,.1) coordinate (C2L1);
}
\newcommand{\tikzrfalayerC}[1]{
    \coordinate (left#1) at (0,0);
    \coordinate (right#1) at (.2,0);
    \draw[help lines,step=.2,shift={(0,.1)}] (0,-0.6) grid (0.2,0.4);
    \draw[A,line width=.6pt]
      (0,-.3) coordinate (A1L2) rectangle (.2,.3) coordinate (A2L2);
    \draw[B,line width=.6pt]
      (0,-.1) coordinate (B1L2) rectangle (.2,.1) coordinate (B2L2);
}
\newcommand{\tikzrfalayerD}[1]{
    \coordinate (left#1) at (0,0);
    \coordinate (right#1) at (.2,0);
    \draw[help lines,step=.2,shift={(0,.1)}] (0,-0.6) grid (0.2,0.4);
    \draw[A,line width=.6pt]
      (0,-.1) coordinate (A1L3) rectangle (.2,.1) coordinate (A2L3);
}

\newcommand{\tikzrfalayerCAlt}[1]{
    \coordinate (left#1) at (0,0);
    \coordinate (right#1) at (.2,0);
    \draw[help lines,step=.2,shift={(0,.1)}] (0,-1.2) grid (0.2,1.0);
    \draw[A,line width=.6pt]
      (0,-.3) coordinate (A1L2) rectangle (.2,.3) coordinate (A2L2);
    \draw[B,line width=.6pt]
      (0,-.1) coordinate (B1L2) rectangle (.2,.1) coordinate (B2L2);
}

\newcommand{\tikzrfalayerCAltAlt}[1]{
    \coordinate (left#1) at (0,0);
    \coordinate (right#1) at (.2,0);
    \draw[help lines,step=.2,shift={(0,.1)}] (0,-0.6) grid (0.2,0.4);
    \draw[help lines,step=.2,shift={(0,.1)}] (0,-0.6) grid (0.4,0.4);
    \draw[A,line width=.6pt]
      (0,-.3) coordinate (A1L2) rectangle (.2,.3) coordinate (A2L2);
    \draw[B,line width=.6pt]
      (0,-.1) coordinate (B1L2) rectangle (.2,.1) coordinate (B2L2);
}

\newcommand{\tikzrfalayerDAlt}[1]{
    \coordinate (left#1) at (0,0);
    \coordinate (right#1) at (.2,0);
    \draw[help lines,step=.2,shift={(0,.1)}] (0,-0.6) grid (0.2,0.4);
    \draw[A,line width=.6pt]
      (0,-.1) coordinate (A1L3) rectangle (.2,.1) coordinate (A2L3);
}

\begin{tikzpicture}[
    background/.style={fill=white!80!black,pattern=north west lines},
    imagearea/.style={fill=white},
    A/.style={red,fill=white!90!red,fill opacity=.35,even odd rule},
    B/.style={green,fill=white!90!green,fill opacity=.35},
    C/.style={blue,fill=white!90!blue,fill opacity=.35},
    textnode/.style={align=center,shift={(0,-2.3)}}
    ]
    \def\rectsize{2.52}
    \def\imgsize{1.5}
    \def\Asize{2.3}
    \def\Bsize{1.5}
    \def\Csize{.5}

    \begin{scope}[scale=.7,local bounding box=base]
      \tikzrfaimage[scale=.7]
      \node[textnode] {input image\\with padding};
      \begin{scope}[shift={(4,0)}]
        \tikzrfalayerA{0}
        \node[textnode] {input\\layer};    
      \end{scope}
      \begin{scope}[shift={(6,0)}]
        \tikzrfalayerB{1}
        \node[textnode] {utilized\\layer};    
      \end{scope}
      \begin{scope}[shift={(8,0)}]
        \tikzrfalayerC{2}     
        \node[textnode] {utilized\\layer};
      \end{scope}
      \begin{scope}[shift={(10,0)}]
        \tikzrfalayerD{3}
        \node[textnode] {unutilized\\layer};
      \end{scope}

      \draw[A,dotted,line width=1.pt] (A2) -- (A2-|left0) (A1-|A2) -- (A1-|left0);
      \draw[A] (A1-|right0) -- (A1L1-|left1) (A2-|right0) -- (A2L1-|left1);
      \draw[A] (A1L1-|right1) -- (A1L2-|left2) (A2L1-|right1) -- (A2L2-|left2);
      \draw[A] (A1L2-|right2) -- (A1L3-|left3) (A2L2-|right2) -- (A2L3-|left3);

      \draw[B,dotted,line width=1.pt] (B2) -- (B2-|left0) (B1-|B2) -- (B1-|left0);
      \draw[B] (B1-|right0) -- (B1L1-|left1) (B2-|right0) -- (B2L1-|left1);
      \draw[B] (B1L1-|right1) -- (B1L2-|left2) (B2L1-|right1) -- (B2L2-|left2);

      \draw[C,dotted,line width=1.pt] (C2) -- (C2-|left0) (C1-|C2) -- (C1-|left0);
      \draw[C] (C1-|right0) -- (C1L1-|left1) (C2-|right0) -- (C2L1-|left1);
    \end{scope}
    \begin{scope}[scale=.4,shift={(-6,-11)},local bounding box=opt1,
        A/.style={red, fill=white!90!green,fill opacity=.3,even odd rule},
        B/.style={green,fill=white!90!blue,fill opacity=.3},
      ]
      \def\imgsize{1.5}
      \def\Asize{1.5}
      \def\Bsize{.85}

      \tikzrfaimage[scale=.4]
      \begin{scope}[shift={(4,0)}]
        \tikzrfalayerA{0}
        \path node[below, yshift=-1.25cm, align=center] {All layers are now fully utilized\\(better predictive performance)};

      \end{scope}
      \begin{scope}[shift={(6,0)}]
        \tikzrfalayerB{1}

      \end{scope}
      \begin{scope}[shift={(8,0)}]
        \tikzrfalayerCAlt{2}     
      \end{scope}
      \begin{scope}[shift={(10,0)}]
        \tikzrfalayerDAlt{3}
      \end{scope}

      \draw[A,dotted,line width=1.pt] (A2) -- (A2-|left0) (A1-|A2) -- (A1-|left0);
      \draw[A] (A1-|right0) -- (A1L1-|left1) (A2-|right0) -- (A2L1-|left1);
      \draw[A] (A1L1-|right1) -- (A1L2-|left2) (A2L1-|right1) -- (A2L2-|left2);
      \draw[A] (A1L2-|right2) -- (A1L3-|left3) (A2L2-|right2) -- (A2L3-|left3);

      \draw[B,dotted,line width=1.pt] (B2) -- (B2-|left0) (B1-|B2) -- (B1-|left0);
      \draw[B] (B1-|right0) -- (B1L1-|left1) (B2-|right0) -- (B2L1-|left1);
      \draw[B] (B1L1-|right1) -- (B1L2-|left2) (B2L1-|right1) -- (B2L2-|left2);

      \draw[C,dotted,line width=1.pt] (C2) -- (C2-|left0) (C1-|C2) -- (C1-|left0);
      \draw[C] (C1-|right0) -- (C1L1-|left1) (C2-|right0) -- (C2L1-|left1);
    \end{scope}

    \begin{scope}[scale=.4,shift={(10,-11)},local bounding box=opt2,
      ]
      \tikzrfaimage[scale=.4]
      \begin{scope}[shift={(4,0)}]
        \tikzrfalayerA{0}
        \path node[below, yshift=-1.25cm, align=center] {Only fully utilized layers remain\\(better predictive performance)};

      \end{scope}
      \begin{scope}[shift={(6,0)}]
        \tikzrfalayerB{1}

      \end{scope}
      \begin{scope}[shift={(8,0)}]
        \tikzrfalayerCAltAlt{2}     
      \end{scope}
      \begin{scope}[shift={(10,0)},opacity=.2]
        \tikzrfalayerD{3}
      \end{scope}

      \draw[A,dotted,line width=1.pt] (A2) -- (A2-|left0) (A1-|A2) -- (A1-|left0);
      \draw[A] (A1-|right0) -- (A1L1-|left1) (A2-|right0) -- (A2L1-|left1);
      \draw[A] (A1L1-|right1) -- (A1L2-|left2) (A2L1-|right1) -- (A2L2-|left2);
      \draw[A] (A1L2-|right2) -- (A1L3-|left3) (A2L2-|right2) -- (A2L3-|left3);

      \draw[B,dotted,line width=1.pt] (B2) -- (B2-|left0) (B1-|B2) -- (B1-|left0);
      \draw[B] (B1-|right0) -- (B1L1-|left1) (B2-|right0) -- (B2L1-|left1);
      \draw[B] (B1L1-|right1) -- (B1L2-|left2) (B2L1-|right1) -- (B2L2-|left2);

      \draw[C,dotted,line width=1.pt] (C2) -- (C2-|left0) (C1-|C2) -- (C1-|left0);
      \draw[C] (C1-|right0) -- (C1L1-|left1) (C2-|right0) -- (C2L1-|left1);

      \draw[line width=1ex, red,shift={(10.1,0)},scale=.6]
        (-1,-1) -- (1,1) (-1,1) -- (1,-1);
    \end{scope}

    \draw[->,black,line width=1ex,opacity=.9] (-2.0,0) -|
      node[pos=.7,yshift=-2cm,below, sloped, align=center, rotate=90]{adjust downsampling\\ to slow receptive\\field growth}
      +(-1,-3.3);
    \draw[->,black,line width=1ex,opacity=.9] (7.4,0) -|
      node[pos=.855, yshift=2cm, above, sloped, align=center, rotate=90]{shorter and wider\\architecture}
      +(.5,-3.3);
\end{tikzpicture}